\documentclass{article}

\usepackage[preprint]{neurips_2023}
\usepackage[dvipsnames]{xcolor}
\usepackage{microtype}
\usepackage{graphicx}
\usepackage{booktabs} 
\usepackage{comment}
\usepackage{algorithm}
\usepackage{algpseudocode}
\algtext*{EndWhile}
\algtext*{EndIf}
\algtext*{EndFor}

\usepackage{amsfonts}
\usepackage{bm}
\usepackage{caption}
\usepackage{subcaption}
\usepackage{multirow}
\usepackage{enumitem}
\usepackage{comment}
\usepackage{cuted}
\usepackage{hyperref}
\usepackage{amsmath}
\usepackage{amssymb}
\usepackage{mathtools}
\usepackage{amsthm}

\usepackage[capitalize,noabbrev]{cleveref}
\Crefname{figure}{Fig.}{Figs.}
\Crefname{table}{Tab.}{Tabs.}
\Crefname{section}{Sec.}{Secs.}
\Crefname{equation}{Eq.}{Eqs.}
\Crefname{algorithm}{Alg.}{Algs.}
\newcommand\ours{Structurally Prune Anything}
\newcommand\oursacro{SPA}

\newlist{myitemize}{itemize}{3}
\setlist[myitemize,1]{label=\textbullet,leftmargin=0.5in}

\theoremstyle{plain}

\theoremstyle{definition}

\theoremstyle{remark}

\usepackage[utf8]{inputenc} 
\usepackage[T1]{fontenc}    
\usepackage{hyperref}       
\usepackage{url}            
\usepackage{booktabs}       
\usepackage{amsfonts}       
\usepackage{nicefrac}       
\usepackage{microtype}      
\usepackage{xcolor}         

\title{\ours{}:\\ Any Architecture, Any Framework, Any Time}

\author{%
Xun Wang$^{1}$\thanks{equal contribution} \quad John Rachwan$^{2*}$ \quad Stephan Günnemann$^{23}$ \quad Bertrand Charpentier$^2$ \\ 
$^1$CISPA Helmholtz Center for Information Security \quad $^2$Pruna AI \\
$^3$Department of Computer Science \& Munich Data Science
Institute, Technical University of Munich\\
\texttt{xun.wang@cispa.de}\\
\texttt{\{john.rachwan,stephan.guennemann,bertrand.charpentier\}@pruna.ai}}

\begin{document}

\maketitle

\begin{abstract}
Neural network pruning serves as a critical technique for enhancing the efficiency of deep learning models. Unlike unstructured pruning, which only sets specific parameters to zero, structured pruning eliminates entire channels, thus yielding direct computational and storage benefits. However, the diverse patterns for coupling parameters, such as residual connections and group convolutions, the diverse deep learning frameworks, and the various time stages at which pruning can be performed make existing pruning methods less adaptable to different architectures, frameworks, and pruning criteria. To address this, we introduce \ours{} (\oursacro{}), a versatile structured pruning framework that can prune neural networks with any architecture, from any framework, and at any stage of training. \oursacro{} leverages a standardized computational graph and ONNX representation to prune diverse neural network architectures without the need for manual intervention. \oursacro{} employs a group-level importance estimation method, which groups dependent computational operators, estimates their importance, and prunes unimportant coupled channels. This enables the transfer of various existing pruning criteria into a structured group style.  As a result, \oursacro{} supports pruning at any time, either before training, after training with fine-tuning, or after training without fine-tuning. In the context of the latter, we introduce Optimal Brain SPA (OB\oursacro{}), an algorithm that achieves state-of-the-art pruning results needing neither fine-tuning nor calibration data. In extensive experiments, \oursacro{} shows competitive to state-of-the-art pruning performance across various architectures, from popular frameworks, at different pruning times.
\end{abstract}

\section{Introduction}
\label{sec:intro}
The increasing complexity and scale of deep learning models \cite{He2015DeepRL,Simonyan15,Dosovitskiy2020AnII} have sparked significant research interest in compression methods. Compression methods, like pruning, aim to reduce model size and computational cost in order to increase inference speed, save energy, and enable deployment on computationally limited devices. In particular, pruning methods mostly fall into two main categories: unstructured pruning which involves setting specific parameters to zero while maintaining the overall network structure \cite{LeCun1989OptimalBD,Hassibi1992SecondOD,Dong2017LearningTP,Han2015LearningBW,lee2018snip,frantar2022obc,Xiao2019AutoPruneAN}, and structured pruning  which involves removing entire channels \cite{Li2016PruningFF,He2018FilterPV,He2017ChannelPF,Lin2020HRankFP,Liu2017LearningEC,earlycrop}. While structured pruning advantageously results in direct computational and memory reduction, it is considered a more complex undertaking. Specifically, structured pruning methods often come with three main challenges.

\textbf{Challenge 1:} The first major challenge consists of the difficulty of applying different structured pruning methods to various model architectures. Indeed, structured pruning entails managing the interdependencies between coupled channels in different layers to modify the model structure without breaking the model connectivity (e.g. see residual connection in \cref{fig:grouping}). Hence, when dealing with coupled channels, most of the existing approaches heavily rely on case-by-case analysis of different model architectures. 
\begin{figure*}[t]
    \centering
    \includegraphics[width=.91\textwidth]{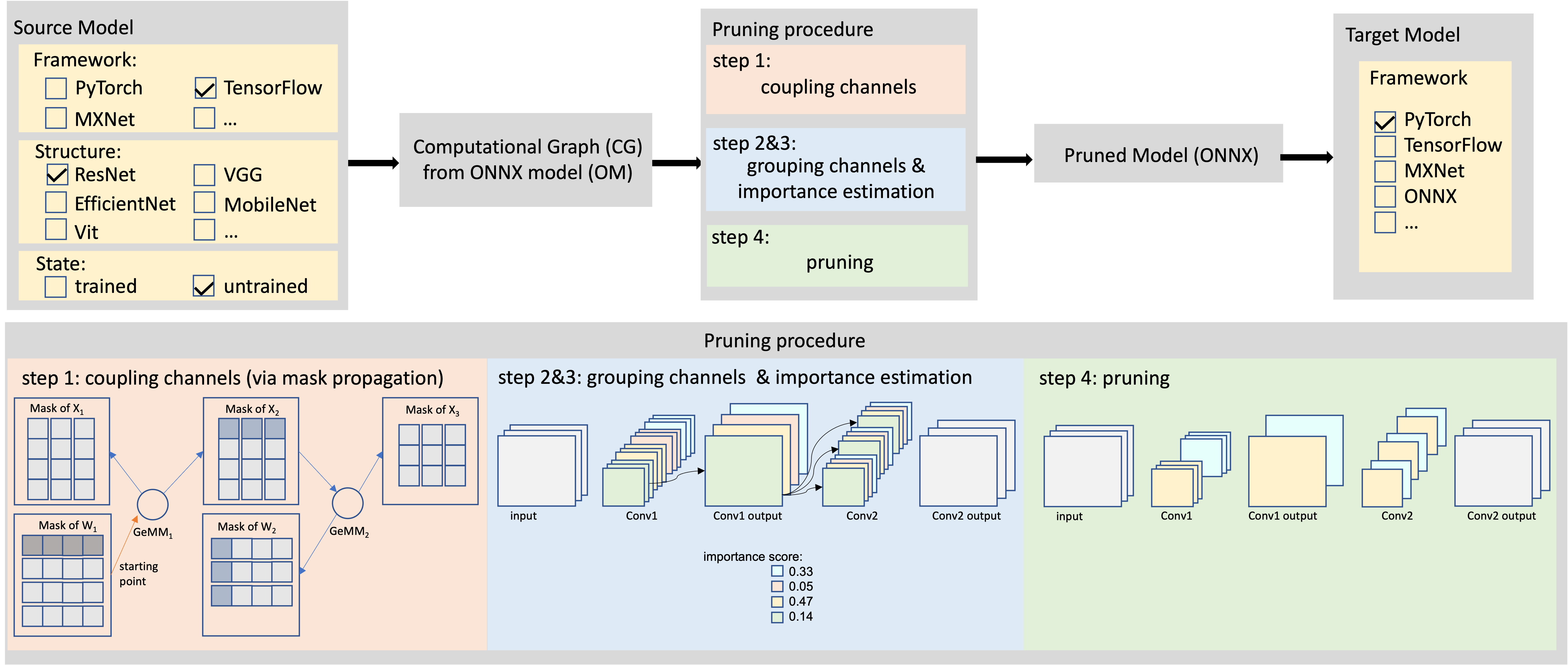}
    \caption{\oursacro{} overview. The source model can be chosen freely from different frameworks with different structures, either trained or not. A computational graph is built to store the dependency information between operators and data. The pruning procedure consists of four steps: coupling channels, grouping channels \& importance estimation, and pruning. After pruning, the pruned model can be converted to other frameworks for further usage.}
    \label{fig:System overview}
    \vspace{-3mm}
\end{figure*}

\textbf{Challenge 2:} The second challenge consists of unifying structured pruning methods in a single framework making pruning possible at any stage of training. Pruning can be done either \emph{before}, \emph{during}, or \emph{after} training. The majority of works adhere to the pruning with fine-tuning approach, which we will refer to as the \emph{train-prune-finetune} setting, and involves conducting finetuning after pruning pre-trained models to restore any performance degradation incurred during the pruning process. Another approach consists in pruning a model before training, which we will refer to as the \emph{prune-train} setting, thus allowing to directly train a sparse model. Nonetheless, a more challenging yet advantageous scenario is the pruning without fine-tuning setting, which we will refer to as the \emph{train-prune} setting \cite{Lazarevich2021PosttrainingDN}, wherein no additional training is permitted after pruning a pre-trained model. Instead, the train-prune setting has only access to a limited set of calibration data \cite{frantar2022obc,Frantar2023SparseGPTML}, or, even more challenging, has access to no calibration data \cite{Srinivas2015DatafreePP} for the pruning step.

\textbf{Challenge 3:} The third challenge is that existing pruning methods are not only often designed with specific architectures or training paradigms in mind, but they are also further entrenched by the deep learning frameworks they were developed. This framework specificity arises due to several factors: differences in computational graph, definition of specific layers, and the existence of unique APIs and optimization libraries. As such, a pruning method effective in one setting may require non-trivial adaptations to be ported to another framework or architecture, complicating its general applicability. Hence, the third challenge lies in crafting an approach robust enough to transcend the limitations imposed by framework-specific constraints and progress toward a unified, generalizable approach to model pruning.

Previous works have tried to address parts of these three challenges. For instance, DepGraph \cite{fang2023depgraph} and OTO-v2 \cite{chen2023otov2} enables the automatic pruning of different networks by maintaining a dependency graph. However, they lack the ability to support models other than PyTorch and only support pruning after training with or without fine-tuning scheme. Further, DFPC \cite{narshana2023dfpc} proposed a method to prune coupled channels data-free without fine-tuning, but it lacks the ability to adapt to different architectures and frameworks. 

To jointly tackle the aforementioned three challenges, we propose \ours{} (\oursacro{}), an architecture-and-framework-agnostic neural network pruning method, which supports different criteria that encompass the previous three settings we defined. We show an overview of our method in \cref{fig:System overview}. Its contributions can be summarized as follows.
    \textbf{(1) Prune Any Framework:} We directly operate on a flexible computational graph compatible across frameworks. To this end, we use the ONNX format. With this procedure, we are the first pruning method that can handle the most common deep learning frameworks.
    \textbf{(2) Prune Any Architecture:} We propose a four-step procedure for the structured pruning of grouped channels. This procedure allows automatic pruning of neural networks with any structures, and the easy transfer of many existing pruning criteria for a grouped structured version, often achieving superior performance/efficiency trade-off.
    \textbf{(3) Prune Any Time:} We propose a group-level importance estimation method, enabling pruning at any training stage including prune-train, train-prune-finetune, and train-prune. In the latter setting, we propose a \emph{novel} method Optimal Brain \oursacro{} (OBSPA) which achieves state-of-the-art results with ResNet50 on CIFAR10 and VGG19 on CIFAR100 without the need for calibration data.
\section{Related Works}
\label{sec:related_works}

\textbf{Pruning criteria}: To determine which connection or neuron should be pruned, various pruning criteria are employed to identify their importance. Most pruning research has followed the approach pioneered by \cite{Han2015LearningBW} of using weight magnitudes as importance scores. These include \cite{Li2016PruningFF,He2018SoftFP}. However, the drawback of only using weight magnitudes is that the network has to be pre-trained in order for it to achieve good performance. Therefore, some approaches have focussed on augmenting them with first-order and second-order information, which allows for the pruning to be applied even on a randomly initialized network \cite{lee2018snip,verdenius2020pruning,Wang2020Picking,earlycrop}. Most recently, due to the rise of generative models and their growing costs, pruning research has shifted its focus towards removing the need to fine-tune after pruning. These approaches generate importance scores by solving complex optimization problems that attempt to preserve the per-layer outputs of the model \cite{frantar2022obc,Frantar2023SparseGPTML}. We recommend interested readers to refer to the following surveys \cite{He2023StructuredPF,Blalock2020WhatIT} for a more comprehensive overview of the previously discussed approaches as well as additional ones such as activation-based \cite{He2017ChannelPF,ThiNet_ICCV17,Lin2020HRankFP,Yu2017NISPPN,Zhuang2018DiscriminationawareCP}, and regularization based \cite{Liu2017learning,zhonghui2019gate,Huang2017DataDrivenSS,ding2021resrep} variants.

\textbf{Pruning coupled channels}: Research on pruning coupled parameters has been a prominent area of focus since the initial stages of structural pruning, with techniques like slimming \cite{Liu2017LearningEC} and ThiNet \cite{ThiNet_ICCV17} aiming to identify and remove such dependencies. However, manually analyzing parameter inter-dependencies can be an exceedingly arduous task, particularly when applied to complex networks such as DenseNet \cite{Huang2016DenselyCC}. Some works have emerged to discover the complex relationships between layers by automatically uncovering the dependencies between the layers. Group Fisher pruning \cite{Liu2021GroupFP} introduces a versatile channel pruning approach applicable to complex structures by building the network's dependency graph. DFPC \cite{narshana2023dfpc} prunes the coupled channels in a one-shot and data-free manner, it introduces the concept of Data Flow Couplings (DFCs).  DFCs are tuples that describe a set of layers and the transformations between them that couple the channels of the output of one layer to the channels of the input of another layer. Most recently, OTO-v2 \cite{chen2023otov2} and DepGraph \cite{fang2023depgraph} also address the problem by building a dependency graph. On one hand, OTO-v2 traces the operator connectivity in CNNs, residual, or transfomer architectures but requires a specific training rountine with Zero-Invariant-Group partitions. On the other hand, DepGraph traces the model's gradient functions in the backward pass to generalize to mutliple architectures such as CNNs, RNNs, GNNs, or transformers. Both OTO-v2 and DepGraph are restricted to Pytorch models and use dependency graphs which capture limited information thus requiring a more manual understanding of some networks like ViT.

\textbf{Pruning time}:
Numerous pruning methods utilize distinct pruning configurations, which exhibit variations in terms of the initial state of the model subjected to pruning (i.e., whether it is a fully trained model or randomly initialized) and the necessity of fine-tuning the pruned model. In this paper, we are mainly interested in the following frameworks: (1) train-prune-finetune \cite{Han2015LearningBW} where a pre-trained model is finetuned after the pruning step, (2) prune-train \cite{lee2018snip,verdenius2020pruning,Wang2020Picking,earlycrop}, where a randomly initialized model is pruned and then trained to convergence and (3) train-prune, where a pre-trained model is pruned without the need for further finetuning \cite{Lazarevich2021PosttrainingDN,frantar2022obc,Srinivas2015DatafreePP,narshana2023dfpc}. Some other interesting proposed frameworks are early pruning, where the model is slightly trained at the beginning, after which it is pruned and further fine-tuned \cite{earlycrop,You2020Drawing}, pruning during training, where the pruning and training steps happen simultaneously \cite{evci2020rigging}, as well as \cite{chen2021otov1,chen2023otov2}, where pruned structures are learned during training.
\section{\ours{}}
\label{sec:method}

\begin{figure*}
    \centering
    \begin{subfigure}[b]{0.7\textwidth}
        \centering
        \includegraphics[width=\textwidth]{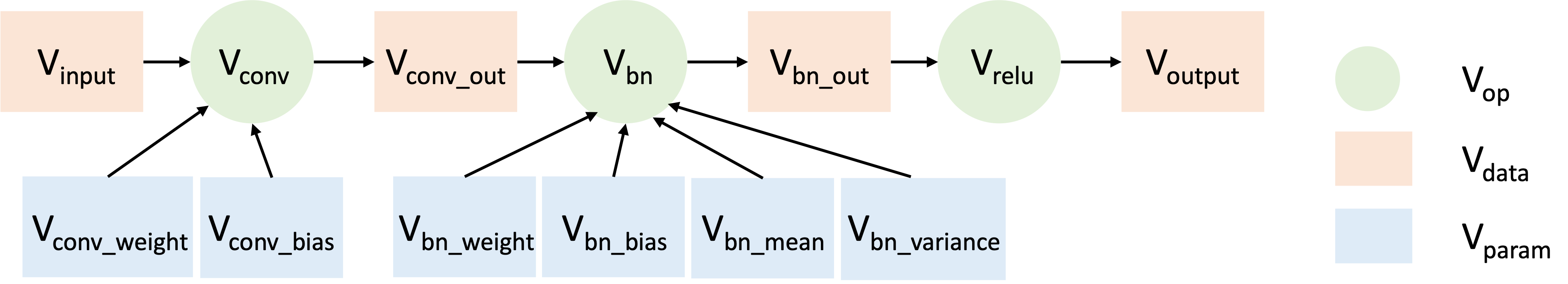}
        \caption{Computational Graph}
        \label{fig:CG}
    \end{subfigure}
    \hfill
    \begin{subfigure}[b]{0.29\textwidth}
        \centering
        \includegraphics[width=\textwidth]{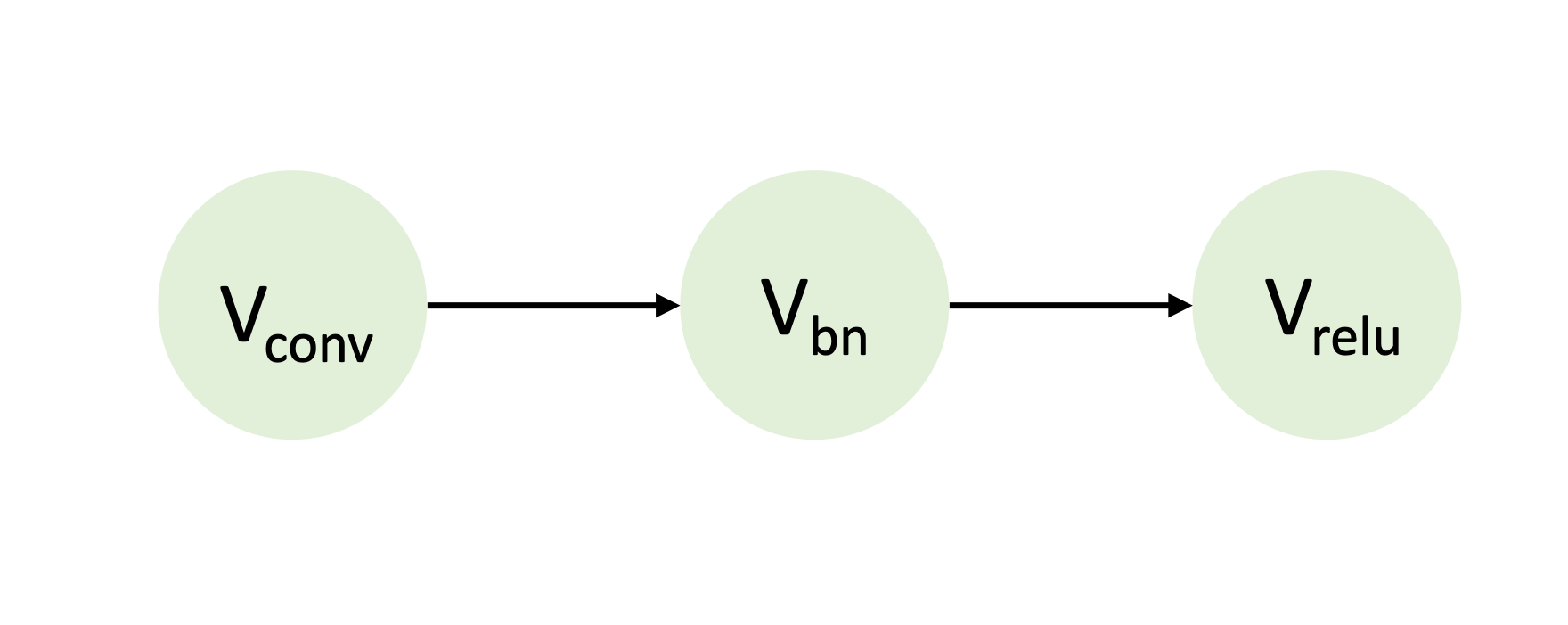}
        \caption{Dependency Graph}
        \label{fig:DG}
    \end{subfigure}
    \caption{Comparison of Computational Graph and Dependency Graph. \cref{fig:CG} is a computational graph. This graph is composed of three operators linked by the data nodes. Convolution and BatchNorm have parameters; they form the parameter nodes in the computational graph. \cref{fig:DG} is the Dependency Graph of the same structure; only information on linked operators is stored.}
    \label{fig:compare_CG_DG}
\end{figure*}

\subsection{Prune Any Framework}
\label{sec:framework}
Our algorithmic analysis critically depends on the computational graph $CG$. For every neural network under consideration, the initial step involves constructing its computational graph. The computational graph is a directed graph that consists of three types of nodes: operator nodes $v_{op}$, which represent basic operators; normal data nodes $v_{data}$, which represent the input and output of operators, and parameter data nodes $v_{param}$ which represent the operators' parameter. Unlike the dependency graph, which only records the dependencies between operators, the computational graph provides essential insights into the relationships among operators and data connections that are necessary to detect dependencies between channels within any model architecture; it meticulously captures crucial information, including the sequencing of operators, the nature of operator-data connections, and the specific data shapes involved. See \cref{fig:compare_CG_DG} for a comparison of the computational graph and dependency graph. 

In our work, we establish a computational graph using the ONNX framework for pruning. The adoption of ONNX offers several notable advantages. First, ONNX provides a static trace of the model, facilitating the straightforward construction of a computational graph based on its explicit representation. Second, ONNX offers a standardized format for model representation. Regardless of how various layers are defined in different frameworks, once converted to ONNX, they assume a uniform sequence of fundamental ONNX operators. This standardization ensures that the analysis of the computational graph remains independent of the underlying frameworks, thus making it framework-agnostic. Third, ONNX enables seamless portability and cross-platform compatibility for models. Models can be effortlessly converted between different frameworks and ONNX. In our work, as depicted in \cref{fig:System overview}, we initially convert models to the ONNX format. This step allows us to construct and examine the computational graph, as well as directly modify the ONNX model. Afterward, we have the option to convert the model back to its original framework.

\subsection{Prune Any Architecture}
\label{sec:architecture}

Given the neural network $f_{\Theta}(x) = y$ where $x$ is the input, $y$ is the predicted output, and $\Theta = \{\theta ^{(1)},..., \theta^{(L)}\}$ are the parameters with $L$ layers, the goal of \oursacro{} is to automatically detect structural correlations within parameters $\theta ^{(1)},..., \theta^{(L)}$, and prune their less important channels or dimensions. To this end, \oursacro{} uses four steps:

\textbf{Step 1: Coupling channels via mask propagation.} Coupling channels are channels that are interconnected due to the dimensional constraints of subsequent layers (e.g. see same colored channels in \cref{fig:System overview,fig:grouping}). Given the computational graph, we employ a mask propagation technique which intuitively aims at finding all the coupled channels for any target channel within any source node. To this end, it initially creates a mask for the target channel in the source node, and iteratively passes it through the operator nodes using predefined rules to identify correlated channels in other parameter nodes. These predefined rules are specific to the standard ONNX operators (see details in \cref{app:mask_propagation}). We explicitly describe the mask propagation algorithm in \cref{alg:op_mask_propagation}. First, it takes as input a computational graph $CG$, a source data node $v_{s}$ and a mask $m_{v_s}$ for the target channel that initializes propagation. Then, it iteratively visits neighboring operator nodes defined by $neighbor(u, CG)$ (see l.5 in \cref{alg:op_mask_propagation}), and propagates masks with the propagation rules defined by $v_{op}.propagate(m_u, u)$ (see l.7 in \cref{alg:op_mask_propagation}).


\begin{algorithm}
\caption{Coupling channels via mask propagation}\label{alg:op_mask_propagation}
 \hspace*{\algorithmicindent} \textbf{Input:} computational graph CG, a source node $s$, a source mask $m_{v_s}$ in which a target channel is masked. \\
 \hspace*{\algorithmicindent} \textbf{Output:} a dict $M$ containing masks in which coupled channels are masked.
\begin{algorithmic}[1]
\State $M = \{v_s: m_{v_s}\}; stack = (v_s, m_{v_s})$
\State \# Visit all correlated data nodes
\While{$stack$}
    \State $u, m_u = stack.pop()$
    \For{$op$ in $neighbor(u, CG)$}
        \State \# Propagate $m_u$ from $u$ via $v_{op}$
        \State $M_{neighbors} = v_{op}.propagate(m_u, u)$
        \For{$v$ in $M_{neighbors}$ not in $M$}
            \State $stack.push(v, M_{neighbors}[v])$
            \State $M.push(v,M_{neighbors}[v]$
            \EndFor
        \EndFor
    \EndWhile
\State return $M$
\end{algorithmic}
\end{algorithm}

\textbf{Step 2: Grouping coupled channels.} After utilizing the mask propagation method to effectively detect coupled channels in the previous step, we now propose to organize them into groups.

We use $G = \{g_1,g_2,..\}$ to denote all groups. A specific group $g_i$ contains a set of coupled channels $CC$ which have the same pattern (e.g. as represented by the group of four colored sets of coupled channels in \cref{fig:grouping}), hence $g_i = \{CC_1,CC_2... \}$. Each coupled channel needs to be deleted as a whole. The individual channels into a given layer in coupled channels are denoted as $C$. Each parameter $\theta$, within a coupled channel, can be assigned an importance score using some score function $S(\theta)$. 


The grouping algorithm is shown in \cref{alg:grouping}. We are given a computational graph, and the algorithm returns all groups. The algorithm loops over all operators in the computational graph to detect coupled channels. To avoid redundant computation, only the output channels of the parameter nodes of each operator are analyzed since the input channels of the operator have been analyzed by its preceding operator. 
\begin{algorithm}
\caption{Grouping coupled channels}\label{alg:grouping}
 \hspace*{\algorithmicindent} \textbf{Input} computational graph $CG$, set $OPS$ with non analyzed operators\\
 \hspace*{\algorithmicindent} \textbf{Output} Groups: $G$ 
\begin{algorithmic}[1]
\State $G \gets \emptyset$
\While{OPS not empty}
        \State $v_{op} = OPS.pop()$
        \State $g = \emptyset; u = parameter\_node(v_{op})$
        \State \# Add all coupled channel $CC$ for group $g$
        \For{$C$ in $u$'s output channels}
            \State $m_u = create\_mask(u, C)$
            \State $CC = coupled\_ch(CG, u, m_u)$  \Comment{\cref{alg:op_mask_propagation}}
            \State $g.add(CC$)
            \EndFor
        \State $G$.insert($g$)
        \State \# Mark visited all analyzed operators in group $g$
        \For{$v_{op}$ in $analyzed\_ops(CG, g)$}
            \State $OPS$.remove($op$)
        \EndFor
    \EndWhile
\State return $G$
\end{algorithmic}
\end{algorithm}


\textbf{Step 3: Importance estimation.} After obtaining the groups, the next step is to assign to each set of coupled channels an importance score which is critical to effectively execute structured pruning. Indeed, this notion has been previously embraced by methodologies such as Group Fisher \cite{Liu2021GroupFP} and DepGraph \cite{fang2023depgraph}, both of which validated its efficacy through empirical experiments. Our approach capitalizes on its inherent autonomous capability to recognize interconnected channels, thereby achieving a higher degree of generality and unity by providing support for a wide range of diverse aggregation and normalization of the individual weight scores. Hence, we propose the following scoring function for the coupled channels $j$ in group $i$:

\begin{equation}
\label{equ:group-level saliency estimation}
s_{i,j}=\underset{CC_l\in g_i}{Norm}(\{AGG(S(\theta_{k}),\forall \theta_k \in CC_j)\})
\end{equation}

The operator $AGG$ aggregates all importance scores $S(\theta_{k})$ within the set of coupled channels $CC_j$ into a singular score which is then normalized over the other coupled channels of the same group via the operator $Norm$ to keep the scores of coupled channels from different groups within the same range for a fair assestment of relative importance. This scoring function is flexible and can encompass different weight scores $S$ (e.g. L1 norm, first-order or second-order, and OBS \cite{Hassibi1992SecondOD} importance scores), different aggregation operators $AGG$ (e.g. mean, max, and product), or different normalization scores $Norm$ (e.g. summation, maximum, or median). The best choice of $AGG$ and $Norm$ function is not fixed over different models; it can be regarded as hyper-parameters that need to be tuned before pruning. We present the detailed algorithm in the Appendix as \cref{alg:group-level importance estimation}.

\textbf{Step 4: Pruning.}
After obtaining the importance score for each set of coupled channels, we simply sort them to identify the least important ones. Subsequently, we locate these channels in the ONNX model, before finally removing them by adjusting the shape and data in the corresponding parameter nodes.

\textbf{Time complexity:} Within a single group $g_i$, we assume that there are $|E_i|$ edges in this sub computational graph and $m_i$ set of coupled channels. The analysis of a single channel takes $\mathcal{O}(|E_i|)$ since the application of the predefined rules takes $\mathcal{O}(1)$ and in the worst case we need to analyze every link between data nodes and operators. If we loop over all channels within one group as suggested in \cref{alg:grouping}, it takes $\mathcal{O}(|E_i|\cdot m_i)$. However, a single mask propagation analysis per group is sufficient because all coupled channels within a group adhere to the same pattern. This reduces the complexity of analyzing one group to $\mathcal{O}(|E_i|)$. For the whole neural network, the analysis in each group is non-overlapping, so the overall complexity of grouping a neural network will still be $\mathcal{O}(|E|)$ where $|E| = \sum{|E_i|}$ is the number of edges of the network. The pruning procedure is simply a loop over all operators which takes $\mathcal{O}(|V_{param}|)$ where $|V_{param}|$ is the total operator number. The overall complexity of our pruning procedure is  $\mathcal{O}(|E|+|V_{param}|)$.

\subsection{Prune Any Time}

In the previous sections, we developed the general \oursacro{} framework to automatically detect coupled channels and assign them an importance score. Leveraging the grouping analysis capabilities of \oursacro{}, we can incorporate many importance estimation criteria (denoted by $S(.)$ in \cref{equ:group-level saliency estimation}) into our framework. These pruning criteria are usually designed to be used at different training stages like in the train-prune-finetune, the purne-train, and the train-prune settings. Beyond enabling the application of pruning at different training stages, the \oursacro{} framework allows to transfer existing pruning criteria into a group-level structured version.

\textbf{Train-Prune-Finetune.} We support criteria that follow the train-prune-finetune scheme. The Magnitude-based criterion is the simplest method to determine a parameter's importance after training. By aggregating the L1-norm following \cref{equ:group-level saliency estimation}, we have \oursacro{}-L1, a group-structured pruning criterion after training. Although the ONNX model is perfect for performing forward passes and building a standardized computational graph, it is not suitable for backward pass for the subsequent fintuning. In order to support the train-prune-finetune scheme, more specifically the finetuning phase, we need to convert the pruned ONNX model to any framework that supports gradient calculation, in our case, we choose PyTorch.

\textbf{Prune-Train.}
We also support the prune-train scheme by applying the same group extension to before-training criteria; for example, we implement \oursacro{}-SNIP, \oursacro{}-Crop and \oursacro{}-GraSP which serve as group-based extensions of the three pre-training pruning criteria, SNIP \cite{lee2018snip}, CroP \cite{earlycrop} and GraSP \cite{Wang2020Picking}, respectively. Those three methods require the calculation of first or second-order derivatives of the parameters which is not natively supported by ONNX. To support gradient-based importance scores, \oursacro{} proposes to convert back the ONNX model into a framework supporting gradient computation like Pytorch. Thus, while \oursacro{} conveniently benefits from the computational graph from ONNX to achieve its framework and architecture agnostic properties (see \cref{sec:framework,sec:architecture}), it also benefits from the practical gradient computations capacities from Pytorch. It is worth mentioning that the conversion between PyTorch and ONNX produces very limited computation overhead, which takes only seconds (see \cref{tab:model_conversion_time}).

\textbf{Train-Prune.}
For the more challenging pruning without fine-tuning setting, we propose a new algorithm, Optimal Brain \oursacro{} (OB\oursacro{}). We leverage the layer-wise sparsification operated by the unstructured pruning methods OBC \cite{frantar2022obc} and its scalable version \cite{Frantar2023SparseGPTML}, to create a novel structured pruning method which can be integrated into our \oursacro{} framework. In the original OBC method, the goal is to find a mask $\bm{M}$ and an updated weight matrix $\bm{\widehat{\Theta}}$ that best preserves the output of each layer given some calibration data $\bm{X}$ and the original weight matrix $\bm{\Theta}$, i.e:

\begin{equation}
\label{equ:layerwise}
    argmin_{\bm{M},\bm{\widehat{\Theta}}}||\bm{\Theta}\bm{X}-(\bm{M}\odot\bm{\widehat{\Theta}})\bm{X}||_2^2
\end{equation}

Based on \cite{Frantar2023SparseGPTML}, the mask $\bm{M}$ is determined according to the layer-OBS score (see \cref{equ:criterion_layer-OBS}), and the weight matrix is updated based on the inverse Hessian $\bm{H^{-1}}=(\bm{X}\bm{X}^T+\lambda\bm{I})^{-1}$. 

Different from OBC, which uses masks with scattered zeros to facilitate unstructured pruning, OB\oursacro{} applies group-level importance estimation to obtain masks that have zeros of entire channels. While the weight updating procedure in OB\oursacro{} is similar to \cite{Frantar2023SparseGPTML}, a crucial difference is that we need to structurally score each coupled channel as a whole with \cref{equ:group-level saliency estimation} to properly delete them without breaking the computational graph (see \cref{fig:compare_SparseGPT_SPA-OBC}). Hence, in contrast with OBC, OB\oursacro{} can deliver real-world efficiency gains on GPU hardware.

Finally, a notable advancement of OB\oursacro{} compared to OBC pertains to the selection of calibration data employed for Hessian computation. \cite{Frantar2023SparseGPTML} use In Distribution (ID) data directly sampled from the training set. However, since the calibration data is only used to preserve the functionality of each layer, we made some relaxations on the previous setting to make it a data-free approach. In a more lenient data-free context, we lack access to the original training data but can employ data from Out-of-Distribution (OOD) sources. The most rigorous data-free scenario entails a lack of access to both ID and OOD data. Calibration samples are drawn from a uniform distribution in this "DataFree" setting. We evaluate both data-driven and data-free approaches in the experiment. Additionally, we propose a batch norm calibration method to improve the performance under ID and OOD settings (see \cref{sec:setting_details} for details). 
\section{Experiments}


In this section, we show that \oursacro{} can prune any framework (see \cref{sec:exp_framework}), any architecture (see \cref{sec:exp_architecture}), any time (see \cref{sec:exp_time}).

\textbf{Dataset.} This work mainly focuses on image classification tasks. We conduct extensive experiments with various datasets including CIFAR-10 \citep{CIFAR10}, CIFAR-100 \citep{CIFAR100}, ImageNette \cite{imagewang} and ImageNet-1k \cite{deng2009imagenet}. We also conduct experiments on text tasks and conduct experiments with SST-2 \cite{SocherEtAl2013:RNTN} dataset, which is a sentiment classification task in NLP. 

\textbf{Evaluation metrics.} The metric employed to evaluate the extent of performance preservation after pruning is classification accuracy. Similarly to \cite{fang2023depgraph,narshana2023dfpc}, our evaluation of efficiency encompasses two primary measures: reduction in floating point operations (FLOPs), denoted as $RF$, and reduction in parameters, denoted as $RP$. It is important to emphasize that the $RF$ metric carries greater significance, as it accurately reflects the actual reduction in computational workload. We employ the fraction of reduced FLOPs and the fraction of reduced parameters, which range from 0 to 1, to facilitate the visualization of these metrics in the figures.

\subsection{Prune Any Framework}
\label{sec:exp_framework}
To validate that \oursacro{} is framework-agnostic, we investigated the pruning of ResNet-18 models derived from PyTorch, TensorFlow, MXNet, and Jax, using the ImageNette dataset as a benchmark for performance evaluation. Models were first initialized and trained within their respective frameworks, after which they were converted to the ONNX format, a reduction of approximately 2$\times$ in FLOPs utilization is targeted after pruning. In addition to the pruning outcomes, we also test the computational overhead incurred during the framework conversion process, all conversions can be completed within seconds (see \cref{tab:model_conversion_time} in Appendix).

\underline{\textit{Observations:}} In \cref{tab:framework-agnostic}, the outcomes of pruning ResNet-18 models from diverse source frameworks are presented. We show that we successfully prune models from all four frameworks, this validates the framework-agnostic prowess of \oursacro{}. The experiment underscores that a model can be successfully converted to the ONNX format in seconds, and then pruned using \oursacro{} framework.

\begin{table}[htpb]
  \caption{Structure pruning with \oursacro{} from $4$ important Deep Learning frameworks with ResNet-18 on ImageNette}\label{tab:framework-agnostic}
  \centering
  \begin{tabular}{l l l l l}
    \toprule
      Framework & ori acc. & pruned acc. & RF & RP \\
    \midrule
      PyTorch & 83.11\% & 82.96\% & 2.16$\times$ & 2.05$\times$\\
      TensorFlow & 82.62\% & 84.30\% & 1.94$\times$ & 5.25$\times$ \\
      MXNet & 84.36\% & 82.77\% & 1.83$\times$ & 8.03$\times$\\
      Jax & 84.46\% & 83.33\% & 2.26$\times$ & 3.64$\times$\\
    \bottomrule
  \end{tabular}
\end{table}

\subsection{Prune Any Architecture}
\label{sec:exp_architecture}

To showcase \oursacro{}'s pruning ability across various architectures, we conducted pruning experiments on a range of $11$ architectures including AlexNet, DenseNet-121, EfficientNet-b0, MobileNet-v2, RegNet\_x\_16gf, ResNet-50, Resnext-50\_32x4d, VGG-16, and Wide-ResNet-101\_2, ViT-base-patch16 on image classification task and DistilBERT on sentiment classification task. These architectures demonstrate a variety of building blocks including skip connections, MLP, convolutions, group convolutions, attention mechanisms, batch normalization, and more. The pruning process was executed within the context of the train-prune-finetune setting with the L1-based criterion being used as the designated importance score. In this experiment, we target a reduction of $\sim 2\times$ in FLOPs for all models.

\underline{\textit{Observations:}} The outcomes, as presented in \cref{tab:architecture-agnostic}, underscore the power of \oursacro{} in supporting a wide range of neural network architectures containing all of the aforementioned building blocks. Even with the simple L1-based criterion, the pruned models achieve very competitive performance compared to their dense counterparts.
\begin{table}[htpb]
  \caption{Structured pruning with \oursacro{} on $11$ architectures on CIFAR10 for image classification models and SST-2 for DistilBERT.}\label{tab:architecture-agnostic}
  \centering
  \begin{tabular}{l l l l l}
    \toprule
      Model & ori acc. & pruned acc. & RF & RP \\
    \midrule
      AlexNet & 89.99\%& 89.80\%& 1.98$\times$& 1.46$\times$\\
      DenseNet-121 & 93.30\%& 94.20\%& 2.14$\times$& 2.35$\times$\\
      EfficientNet-b0 & 94.15\%& 92.,06\%& 2.14$\times$& 1.86$\times$\\
      MobileNet-v2 & 92.33\%& 92.54\%& 2.33$\times$& 2.07$\times$\\
      RegNet\_x\_16gf & 93.83\%& 93.75\%& 2.13$\times$& 1.83$\times$\\
      ResNet-50 & 93.26\%& 93.42\%& 2.13$\times$& 1.98$\times$\\
      ResNext-50\_32x4d & 93.95\%& 93.99\%& 2.07$\times$& 2.05$\times$\\
      VGG16 & 93.82\%& 94.06\%& 2.05$\times$& 2.45$\times$\\
      WideResNet-101 & 93.50\%& 93.41\%& 2.00$\times$& 1.88$\times$\\
      ViT-base & 95.35\%& 96.10\%& 2.05$\times$& 1.50$\times$\\
    \midrule
      DistilBERT & 91.06\%& 88.88\%& 2.04$\times$& 1.50$\times$\\
    \bottomrule
  \end{tabular}
  \vspace{-4mm}
\end{table}

\subsection{Prune Any Time}
\label{sec:exp_time}

\begin{figure*}
    \centering
    \begin{subfigure}[b]{0.21\textwidth}
        \centering
        \includegraphics[width=\textwidth]{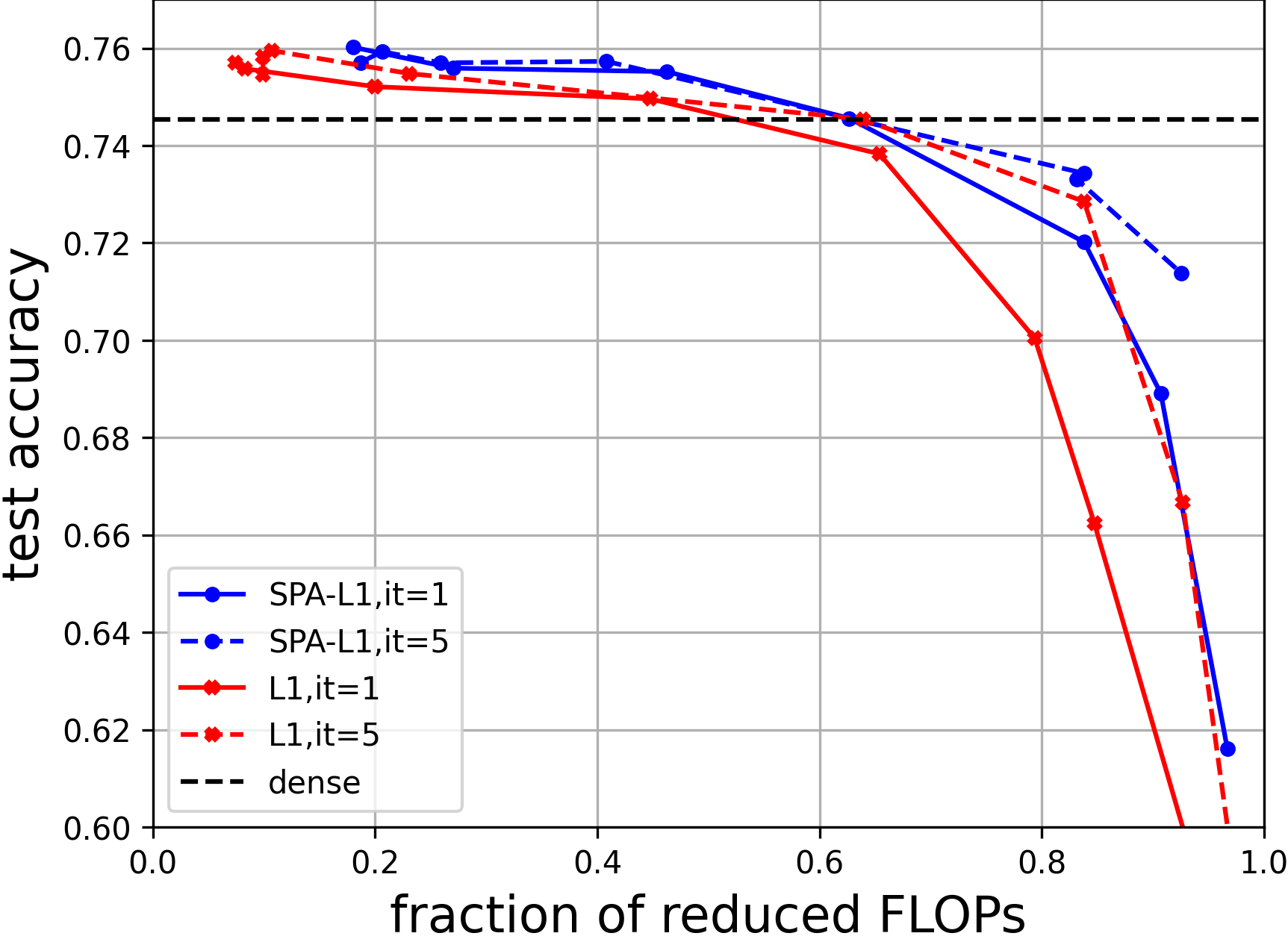}
        \caption{}
        \label{fig:vgg16_L1_RF}
    \end{subfigure}
    \hfill
    \begin{subfigure}[b]{0.21\textwidth}
        \centering
        \includegraphics[width=\textwidth]{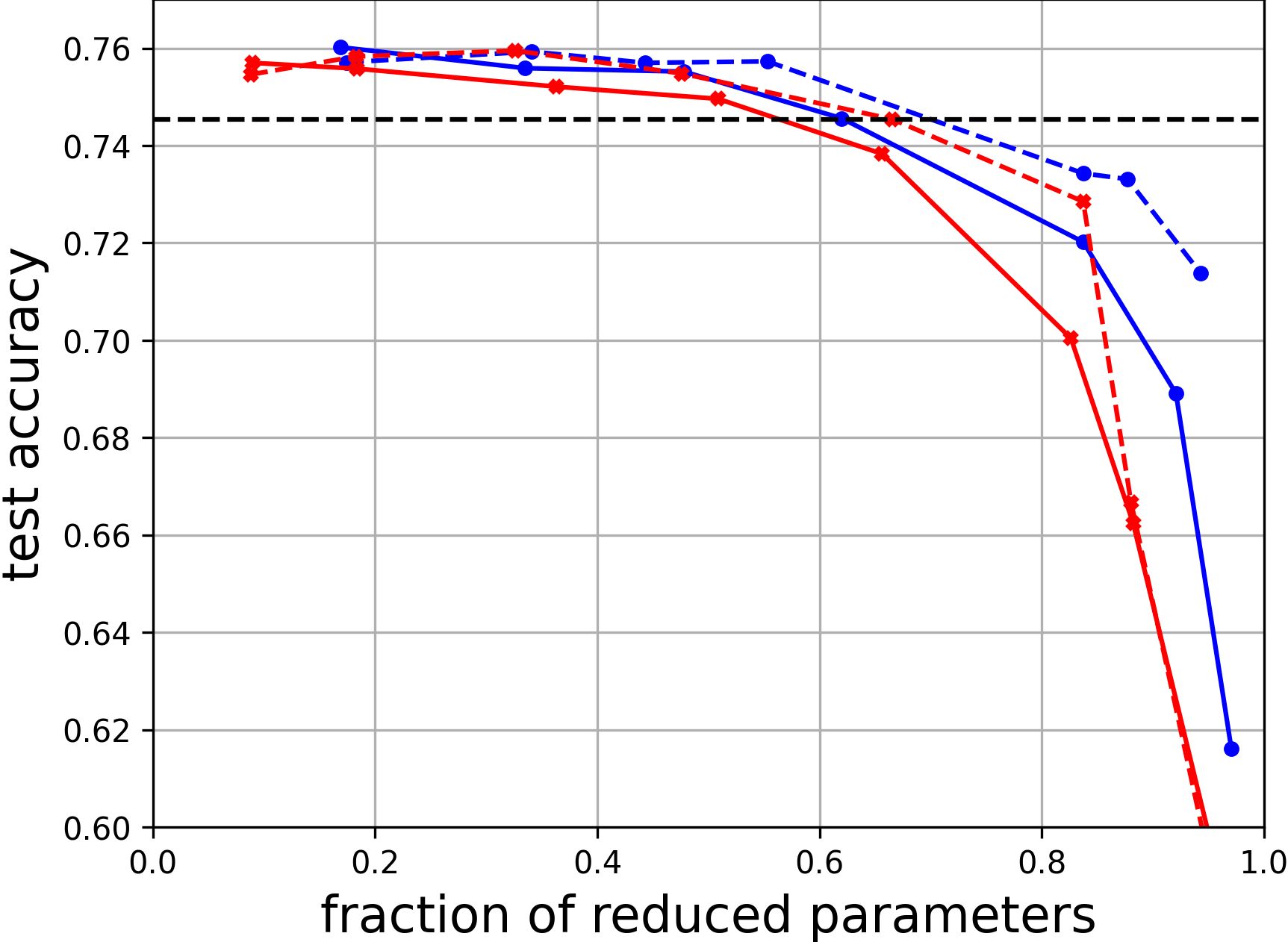}
        \caption{}
        \label{fig:vgg16_L1_RP}
    \end{subfigure}
    \hfill
    \begin{subfigure}[b]{0.21\textwidth}
        \centering
        \includegraphics[width=\textwidth]{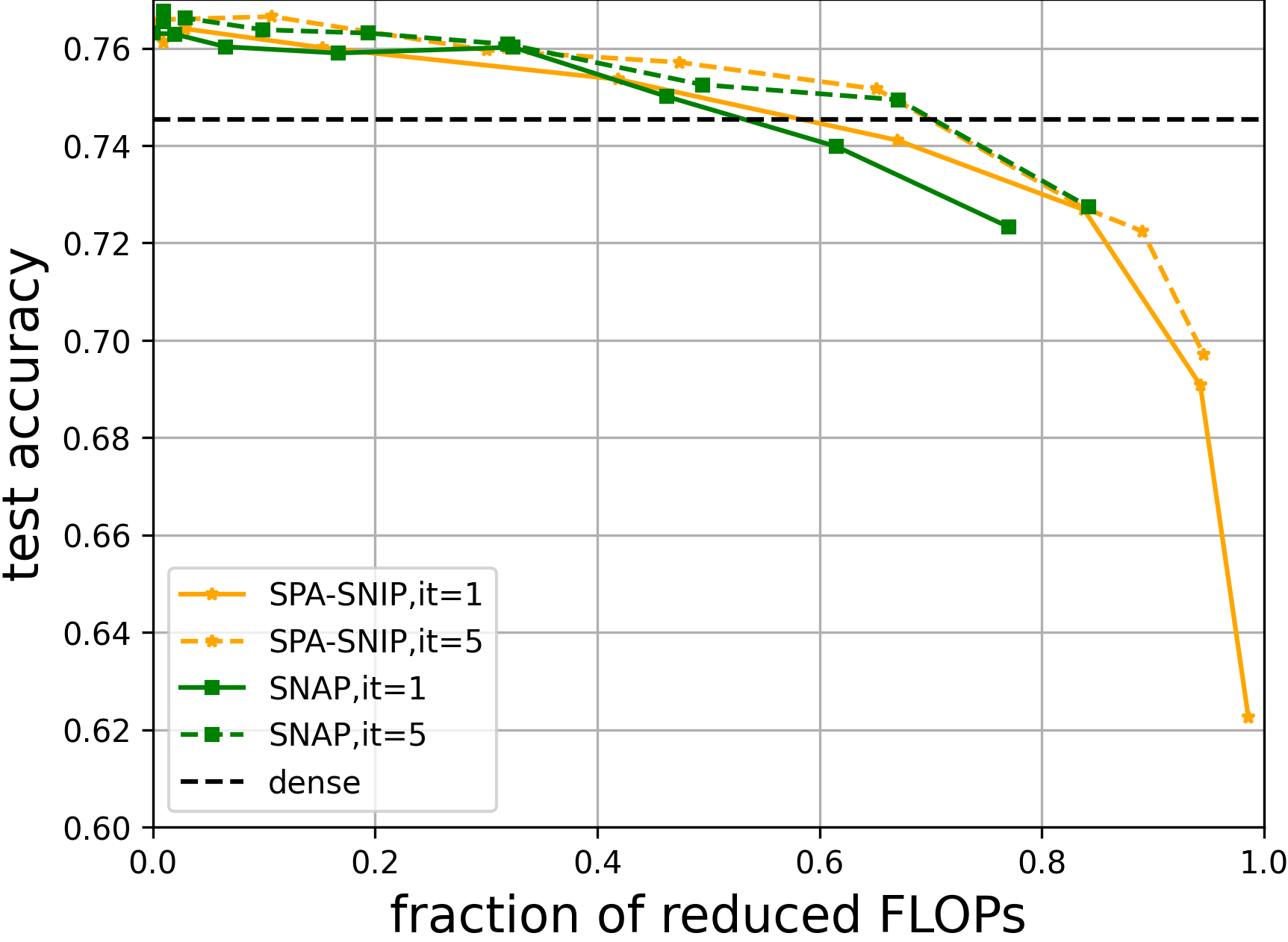}
        \caption{}
        \label{fig:vgg16_SNIP_RF}
    \end{subfigure}
    \hfill
    \begin{subfigure}[b]{0.21\textwidth}
        \centering
        \includegraphics[width=\textwidth]{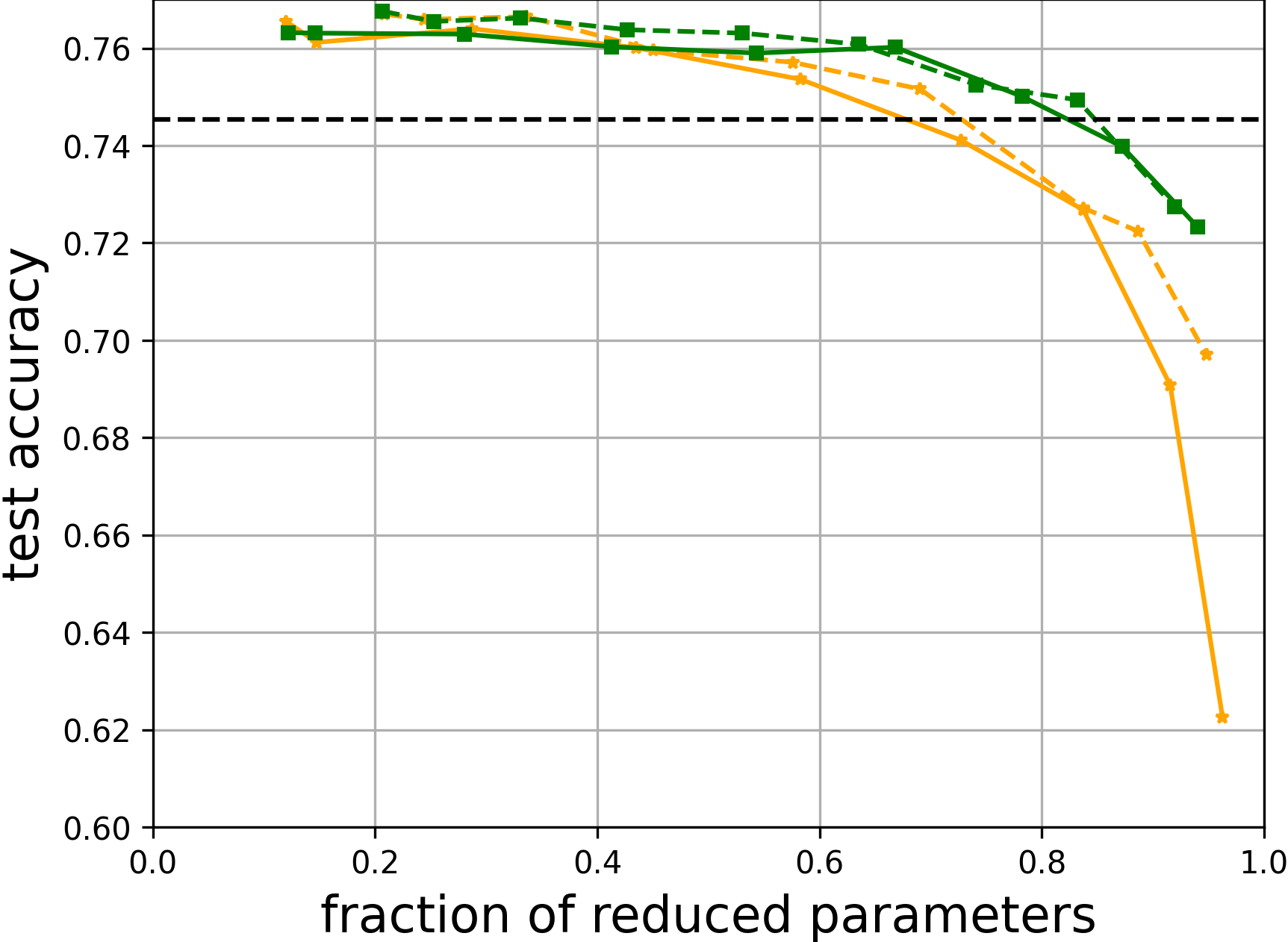}
        \caption{}
        \label{fig:vgg16_SNIP_RP}
    \end{subfigure}
    \newline
    \begin{subfigure}[b]{0.21\textwidth}
        \centering
        \includegraphics[width=\textwidth]{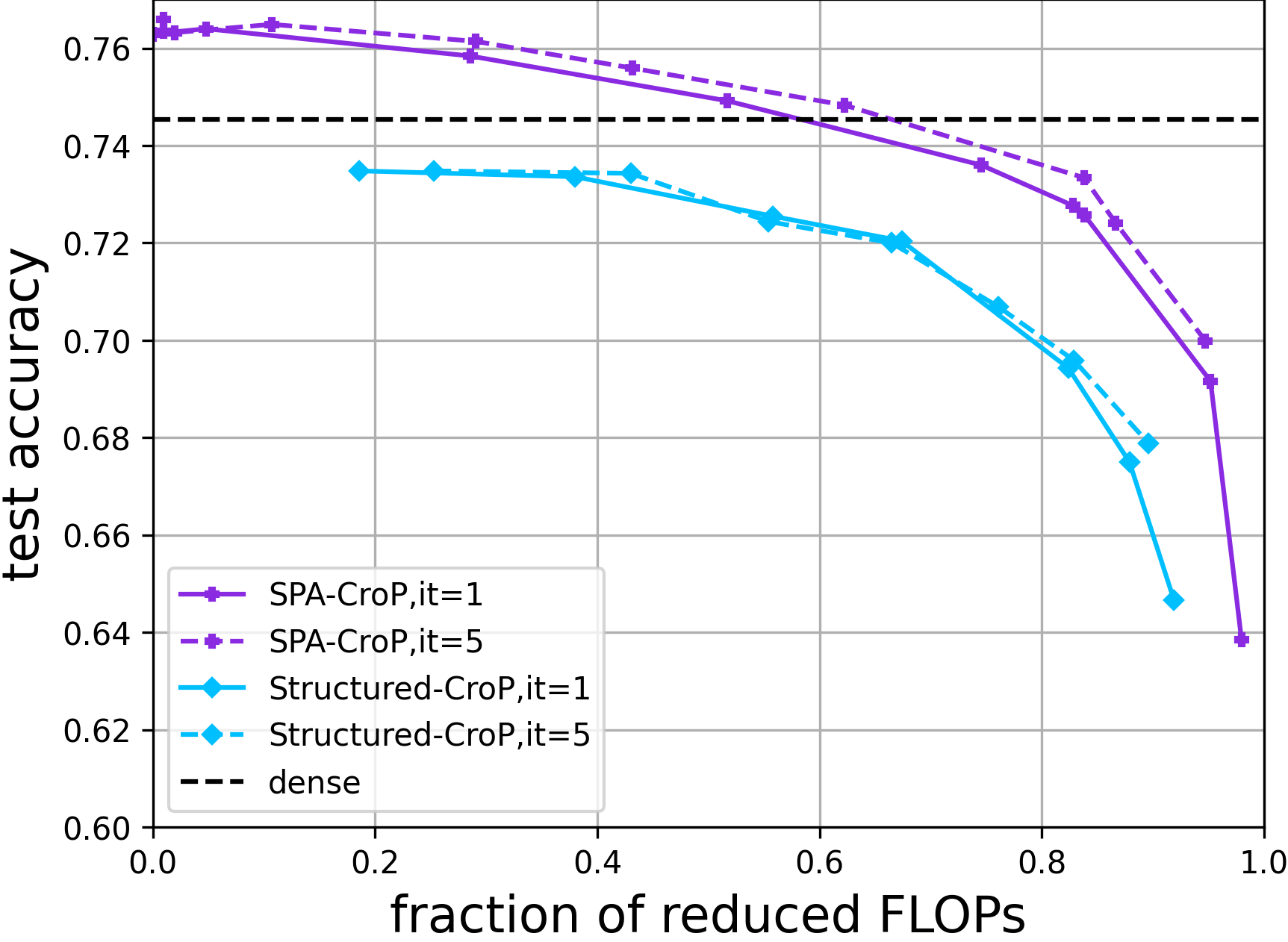}
        \caption{}
        \label{fig:vgg16_CROP_RF}
    \end{subfigure}
    \hfill
    \begin{subfigure}[b]{0.21\textwidth}
        \centering
        \includegraphics[width=\textwidth]{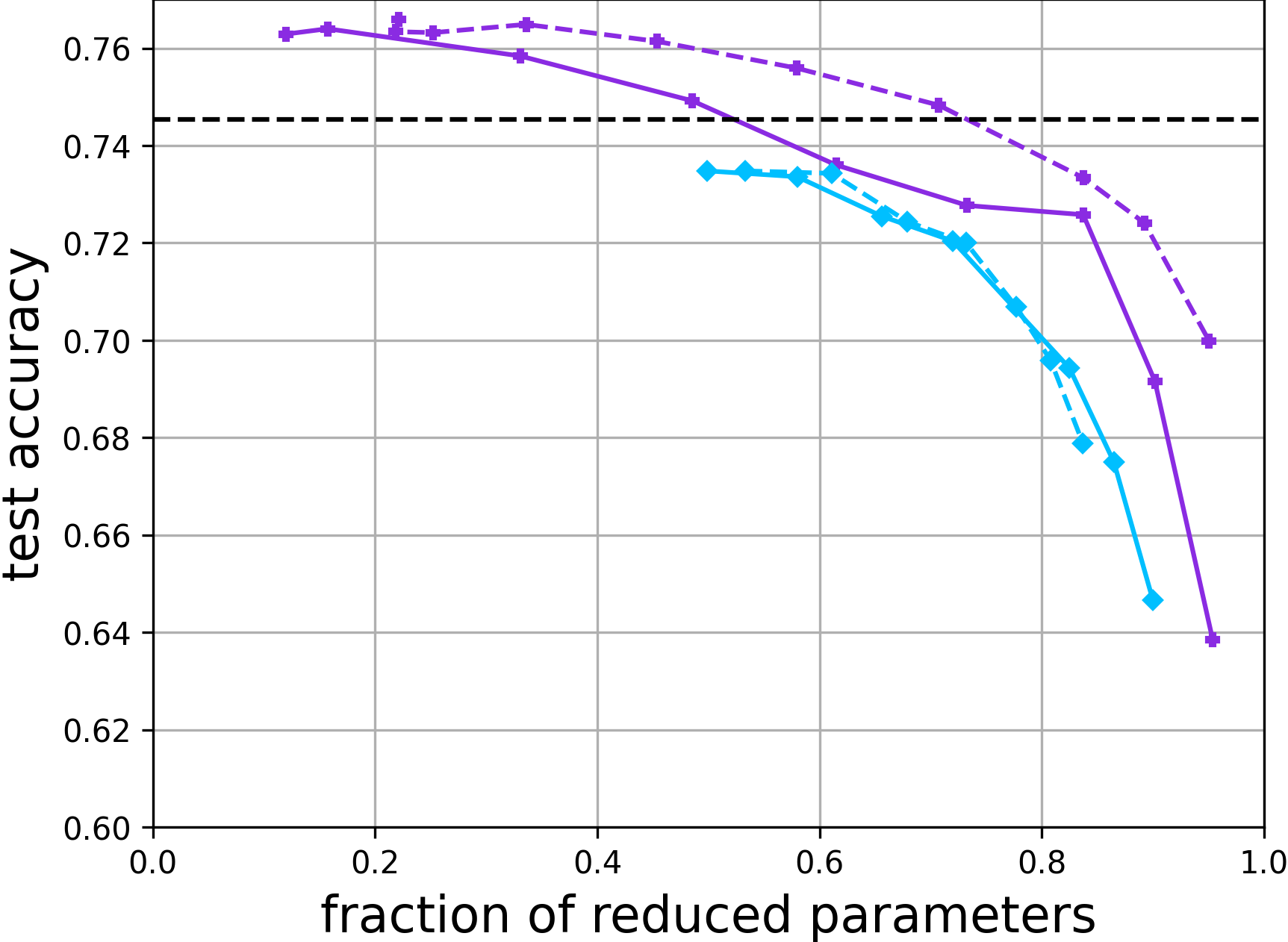}
        \caption{}
        \label{fig:vgg16_CROP_RP}
    \end{subfigure}
    \hfill
    \begin{subfigure}[b]{0.21\textwidth}
        \centering
        \includegraphics[width=\textwidth]{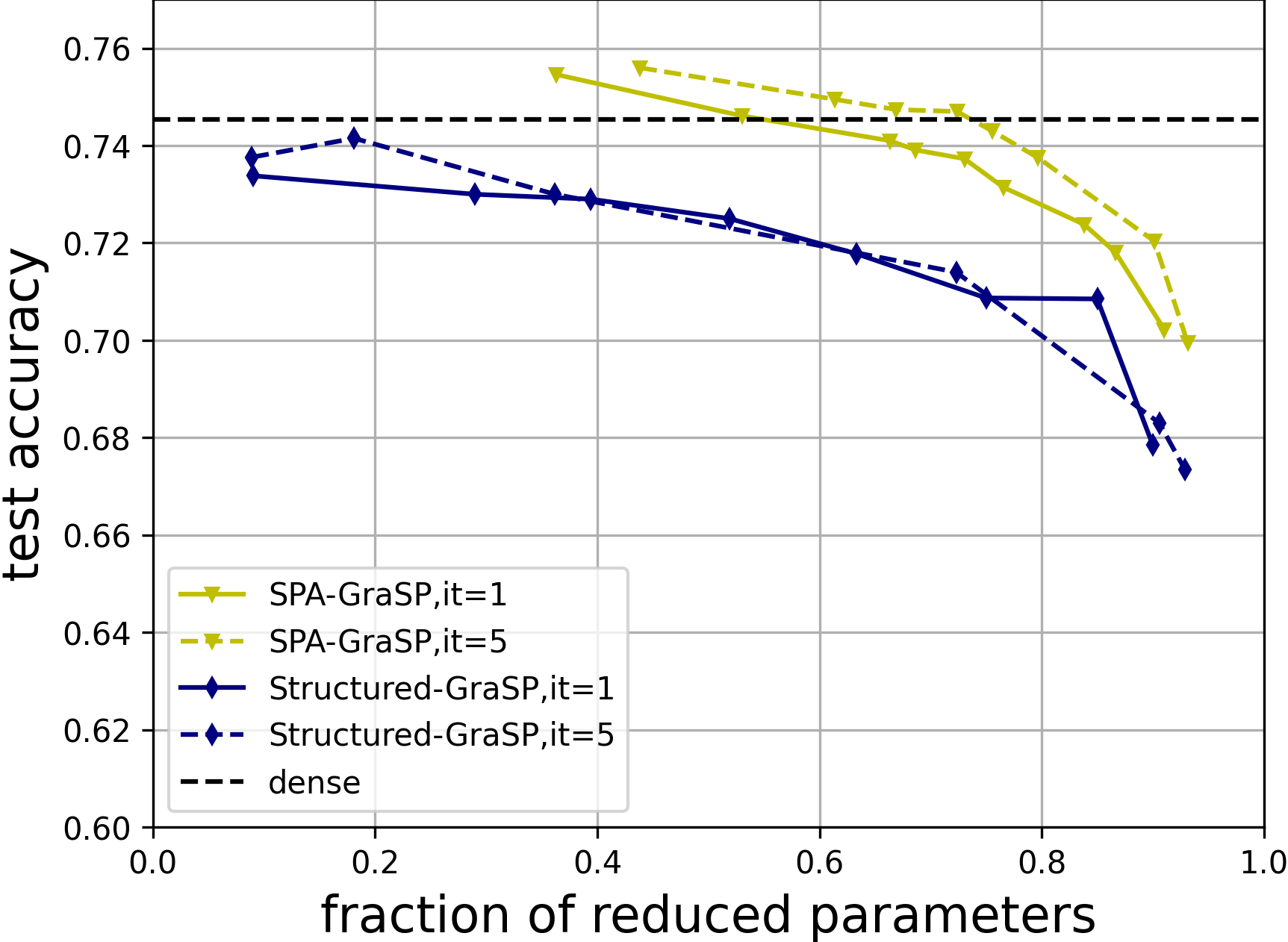}
        \caption{}
        \label{fig:vgg16_GRASP_RF}
    \end{subfigure}
    \hfill
    \begin{subfigure}[b]{0.21\textwidth}
        \centering
        \includegraphics[width=\textwidth]{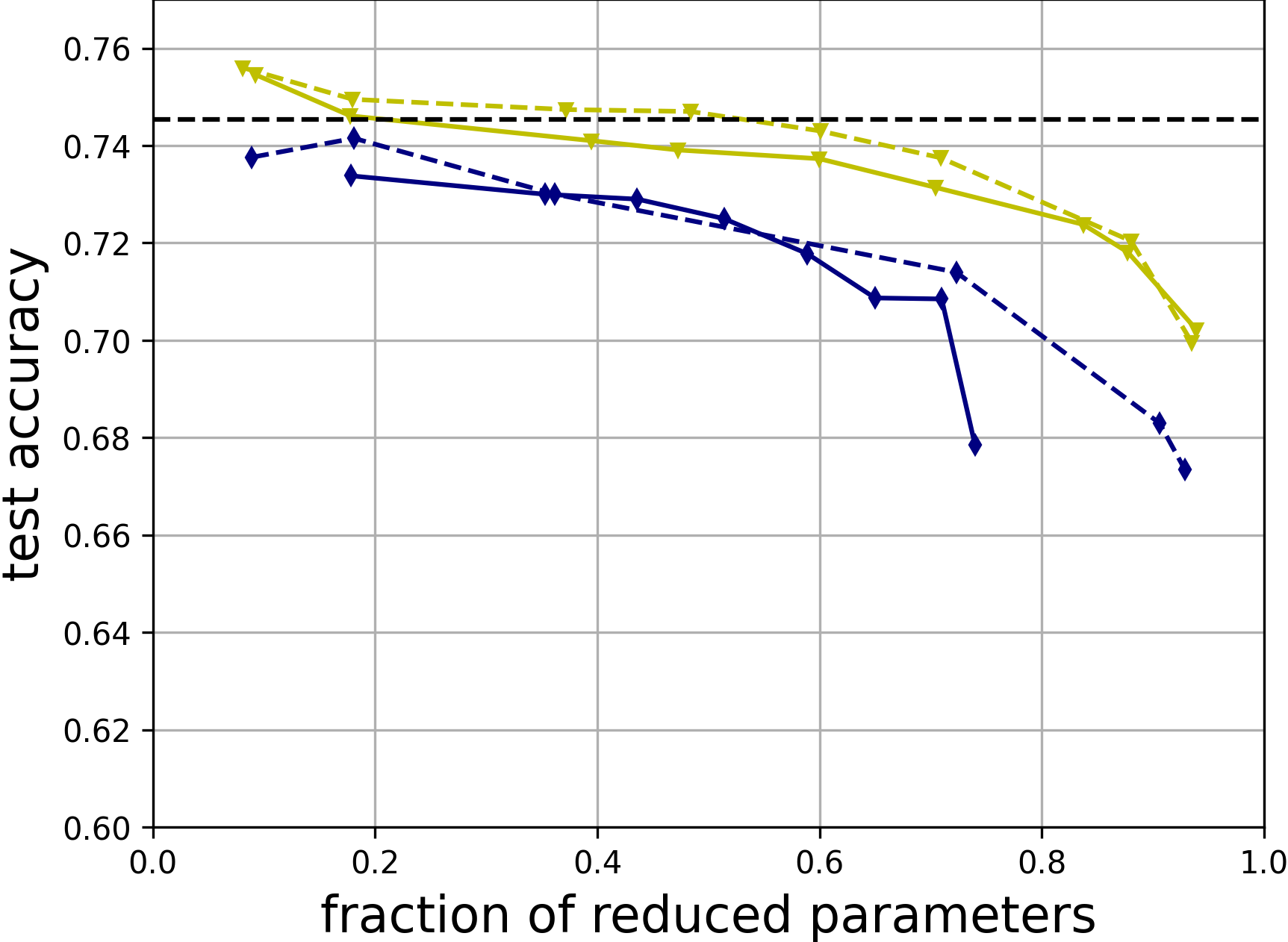}
        \caption{}
        \label{fig:vgg16_GRASP_RP}
    \end{subfigure}
    \vspace{-2mm}
    \caption{Trade off between accuracy and FLOPs/parameters with VGG-16 on CIFAR-100 (\cref{fig:vgg16_L1_RF,fig:vgg16_L1_RP,fig:vgg16_SNIP_RF,fig:vgg16_SNIP_RP,fig:vgg16_CROP_RF,fig:vgg16_CROP_RP,fig:vgg16_GRASP_RF,fig:vgg16_GRASP_RP}). \oursacro{} efficiently implements both the structured and grouped versions of train-prune-finetune criteria like L1 and prune-train criteria like SNAP, CroP and GraSP}
    \label{fig:prune-any-time-cifar100}
    \vspace{-1mm}
\end{figure*}
\begin{table}[htpb]
  \caption{Structured pruning of ResNet-50 on ImageNet with fine-tuning. ”N/R” indicate non-reported results in original papers.}\label{tab:structured-res50-imagenet}
  \centering
  \begin{tabular}{l | c c c c }
    \toprule
     method & top1 acc. &top5 acc. &  RF & RP \\
    \midrule
      Base Model & 76.15\% & 92.86\% & 1$\times$ & 1$\times$  \\
      DFPC \cite{narshana2023dfpc}  & 75.83\% & N/R & 1.98$\times$ & 1.84$\times$\\
      OTO-v2 \cite{chen2023otov2}&75.2\% & 92.2\% & 2.68$\times$ & 2.02$\times$\\
      DepGraph \cite{fang2023depgraph} & 75.83\%& N/R & 2.07$\times$& N/R\\
      \oursacro{}-L1 & 74.83\% & 92.57\% &2.84$\times$ & 2.60$\times$ \\
      \oursacro{}-L1 & 76.39\% &93.29\% &2.18$\times$ & 1.85$\times$ \\
      OB\oursacro{} & 76.59\% & 93.40\%&2.22$\times$ & 1.90$\times$\\
    \bottomrule
  \end{tabular}
\end{table}
\begin{table*}[!htpb]
\vspace{-2mm}
  \caption{Structured pruning of ResNet-50 and VGG-19 on CIFAR-10 and CIFAR-100 without finetuning}\label{tab:structured-oneshot-res50-vgg19}
  \centering
  \resizebox{\textwidth}{!}{
  \begin{tabular}{l | c c c | c c c| c c c | c c c}
    \toprule
      &\multicolumn{6}{c|}{CIFAR-10}  &\multicolumn{6}{c}{CIFAR-100}\\ &\multicolumn{3}{c|}{ResNet-50}  &\multicolumn{3}{c|}{VGG-19} &\multicolumn{3}{c|}{ResNet-50}  & \multicolumn{3}{c}{VGG-19}\\
    method & acc. drop &  RF & RP & acc. drop & RF & RP & acc.drop & RF & RP& acc. drop & RF & RP \\
     
    \midrule
      DFPC & -4.74\% & 1.46 & 2.07 &-3.38\% & 1.68 & 3.16 &-8.53\%& 1.27 & 1.22 & -1.92\%& 1.26 & 1.50 \\
      OB\oursacro{} (ID) & -0.95\% & 1.48 & 1.51 & -0.99\% & 1.71 & 1.44 & -3.73\%& 1.46 & 1.32 & -0.80\%& 1.54 & 1.28\\
      OB\oursacro{} (OOD)  & -1.13\% & 1.48 & 1.52  & -1.67\% & 1.73 & 1.35 & -3.70\%& 1.47 & 1.34 & -1.13\% & 1.54 & 1.28\\
      OB\oursacro{} (DataFree)& -1.34\% & 1.48 & 1.51 & -1.64\% & 1.80 & 1.35& -5.24\%& 1.37 & 1.23 & -1.59\% & 1.47 & 1.28\\
    \bottomrule
  \end{tabular}
  }
\end{table*}

\textbf{Prune with fine-tuning.} By harnessing the channel grouping capability of \oursacro{}, we unlock the potential for extending a multitude of established criteria to a structured group-level pruning paradigm. We aim to underscore the efficacy of our grouped importance estimation method under the pruning with fine-tuning setting on criteria applied both before and after training. We compare the performance of the group L1-based criterion, a train-prune-finetune criterion, to the ungrouped L1 criterion. Then, we delve into the prune-train criteria, where we compare the extended grouping of three prevalent unstructured approaches -- SNIP, CroP, and GraSP -- alongside their structured counterparts, SNAP, structured-CroP, and structured-GraSP. Finally, we also evaluate the OB\oursacro{} with additional fine-tuning.
The postfix "it" denote that pruning is applied in an iterative manner. The evaluation is performed on ResNet-18/CIFAR-10, VGG-16/CIFAR-100, DenseNet-121/ImageNet, ResNet-50/ImageNet and Vit\_b\_16/ImageNet.

\underline{\textit{Observations:}} Through \cref{fig:prune-any-time-cifar100,fig:prune-any-time-cifar10}, we first conclusively showcase \oursacro{}'s versatility in accommodating diverse methodologies. For unstructured criteria like L1-based, SNIP, CroP, and GraSP, the extension to group-structured pruning is easily achieved through the \oursacro{} group analysis. Moreover, we interestingly observe that the performance of the \oursacro{} grouped pruning criteria either matches or outperforms their original structured counterparts. We intuitively explain this observation by the fact that, in contrast with the original structured version of the algorithms, the \oursacro{} grouped versions accounts for all information in a set of coupled channels by aggregating the importance scores over \emph{all} its weights. 
We also observe that gradual iterative pruning consistently yields superior outcomes compared to one-shot channel pruning across nearly all methods. Finally, \oursacro{} matches or outperforms the performance of previous dependency graph approaches on ImageNet in \cref{tab:structured-res50-imagenet,tab:structured-densenet121-imagenet,tab:structured-vit-imagenet}.

\textbf{Prune without fine-tuning.} In this section, our focal point is to showcase the state-of-the-art performance achieved by OB\oursacro{} in the challenging train-prune setting. Following the precedent established by DFPC, we assess the classification performance of pre-trained ResNet-50, ResNet-101, and VGG-19 models on both CIFAR-10 and CIFAR-100 datasets. We also test OB\oursacro{}'s performance on NLP tasks, and compare OB\oursacro{} with L1-based one-shot pruning on pruning a DistilBERT that conducts sentiment classification on SST-2 dataset. Additionally, experiments involving ResNet-50 on the ImageNet dataset have been included in the Appendix (see \cref{app:without-fine-tuning}) to further substantiate our findings.
We conducted experiments in both data-driven and data-free settings. In the experiments, CIFAR-10 serves as OOD dataset for CIFAR-100, and CIFAR-100 serves as OOD dataset for CIFAR-10. We use ax \cite{glue}, another text dataset that contains Natural Language Inference (NLI) problems as OOD dataset for SST-2.

\begin{figure}
    \centering
    \begin{subfigure}[b]{0.3\textwidth}
        \centering
        \includegraphics[width=\textwidth]{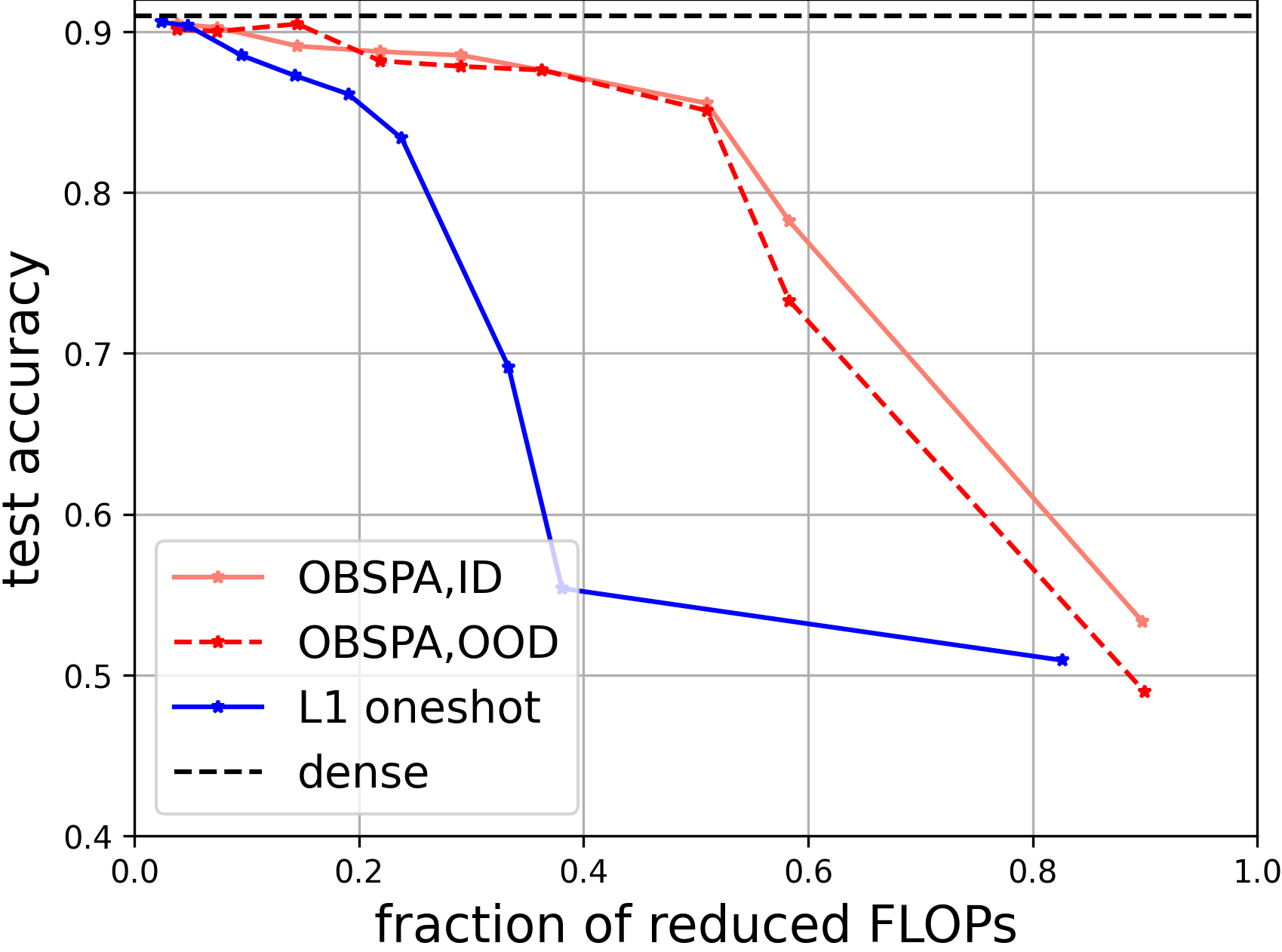}
        \caption{}
        \label{fig:bert_RF}
    \end{subfigure}
    \begin{subfigure}[b]{0.3\textwidth}
        \centering
        \includegraphics[width=\textwidth]{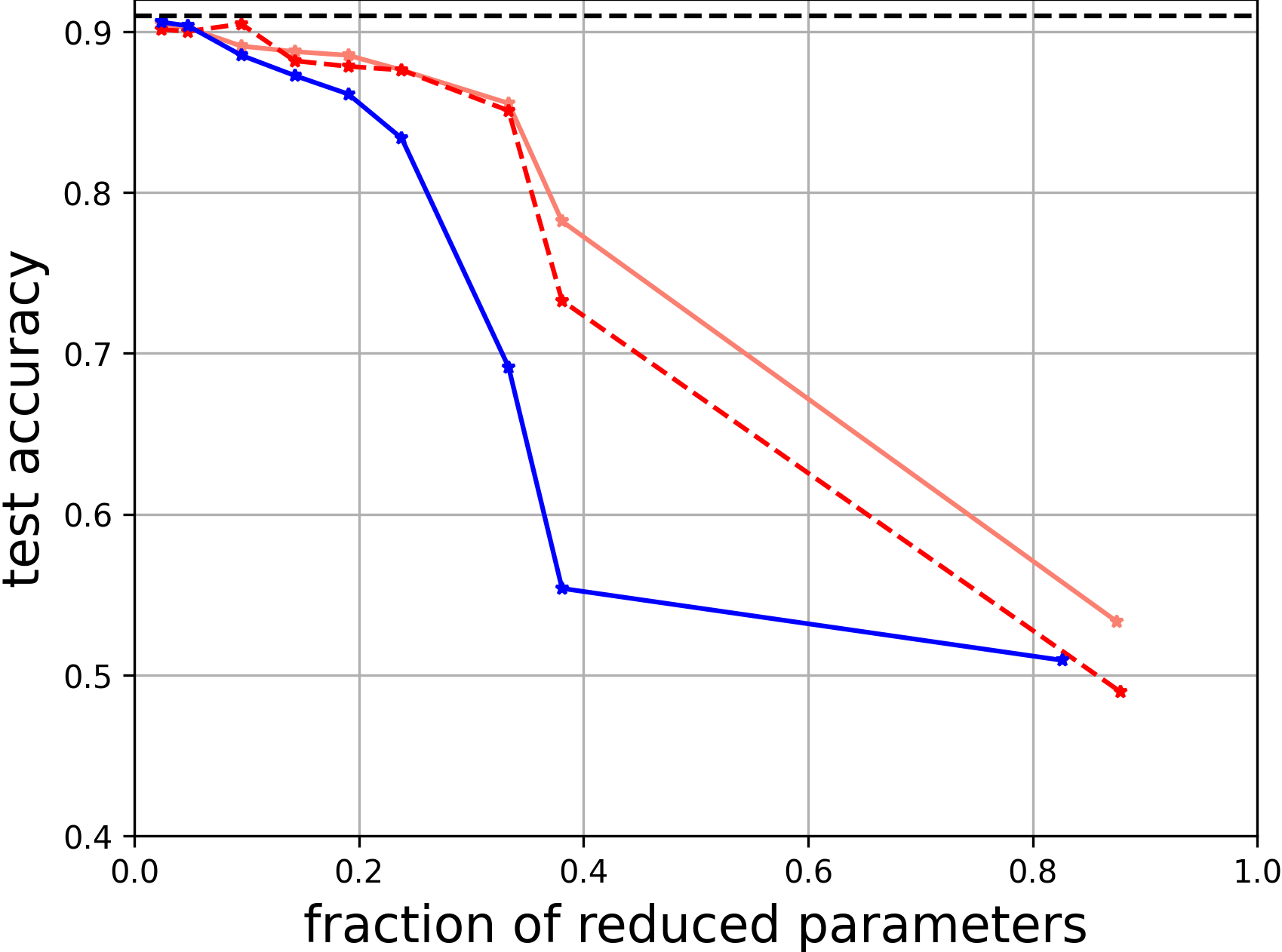}
        \caption{}
        \label{fig:bert_RP}
    \end{subfigure}
    \caption{Trade off between accuracy and FLOPs/parameters with DistilBERT on SST-2 sentiment classificaiton task.}
    \label{fig:bert_SST-2}
    \vspace{-6mm}
\end{figure}

\underline{\textit{Observations:}} We establish a comprehensive comparison between our algorithm and the data-free pruning approach DFPC. \cref{tab:structured-oneshot-res50-vgg19} shows the result of pruning a ResNet-50 and a VGG-19. The outcomes demonstrate the superiority of OB\oursacro{} over DFPC. Specifically, when achieving identical levels of FLOPs reduction, our data-free technique exhibits a mere 1.34\% accuracy drop on the CIFAR-10 classification task with ResNet-50, a remarkable contrast to DFPC's 4.74\% drop. This substantial-performance disparity is also noteworthy on the more complex CIFAR-100 dataset. Notably, for the CIFAR-100 classification with ResNet-50, our data-free approach showcases a 10\% greater FLOPs reduction coupled with a 3.29\% less reduction in accuracy deterioration compared to DFPC. This promising trend is consistently replicated across the ResNet-101 and VGG-19, the ResNet-101 experiment is listed in Appendix. Furthermore, we compare OB\oursacro{} with a basic L1-based one-shot pruning criterion with DistilBERT on SST-2, as suggested in \cref{fig:bert_SST-2}, OB\oursacro{} achieves a much better performance/efficiency trade-off.
Finally, OB\oursacro{} is also much faster than DFPC. We compare the pruning time of our OB\oursacro{} algorithm to DFPC, see results in Appendix \cref{tab:time-structured-oneshot}. We achieved an impressive 6$\times$ speedup for pruning ResNet-50 on both CIFAR and ImageNet-1k dataset.

\section{Conclusion}
In this work, we introduce \oursacro{}, a novel pruning framework that not only automates the pruning of neural networks across diverse architectures but also accommodates models originating from various frameworks. By capitalizing on its inherent capability to 
aggregate interdependent channels, \oursacro{} can convert many pruning criteria into structured pruning algorithms at the group level making it applicable at any time in the training process. Finally, we propose OB\oursacro{}, a structured pruning without fine-tuning algorithm which achieves state-of-the-art performance.

\subsection*{Broader Impact}

This paper presents work that aims to advance the field of efficient Machine Learning (ML). Beyond increasing the speed of ML models, a primary goal of efficiency gains is to reduce the energy and emissions impact of ML applications which is an urgent environmental challenge \cite{Dhar2020TheCI}. Despite the cost reduction that ML compression methods can offer, we encourage practitioners to be aware of the risk of rebound effect and make non-energy policy a standard practice \cite{Dhar2020TheCI}.

\bibliography{ms}
\bibliographystyle{icml2024}

\newpage
\appendix
\section{Method Details}
In this section, we provide a more detailed explanation of our method as well as implementation details.  

\subsection{\oursacro{} group visualization}

\begin{figure*}[t]
    \centering
    \includegraphics[width=\textwidth]{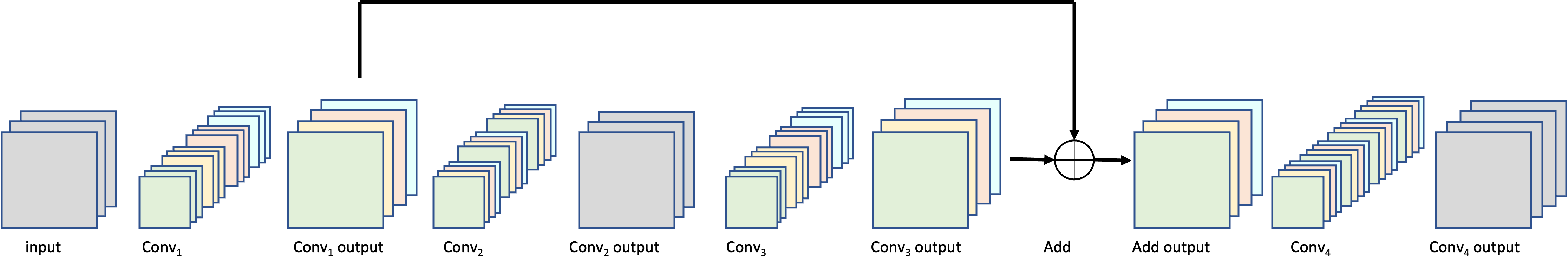}
    \caption{Showcase of a group of a residual structure. Four convolutions with a residual skip form this residual structure. All colored blocks form a group. Within this group, each color represents a coupled channel that must be pruned altogether.}
    \label{fig:grouping}
\end{figure*}

We provide in \cref{fig:grouping} an example of a group of a residual structure with four sets of coupled channels.  Within this group, each color represents a coupled channel that must be pruned altogether.

\subsection{Building Computational Graph}

Starting with an ONNX model, we apply onnx-graphsurgeon, a tool developed in the NVIDIA's TensorRT tookit\footnote{https://github.com/NVIDIA/TensorRT}. This library enables the effortless generation and modification of ONNX models, allowing us to transform the model into a graphsurgeon graph, which we referred to as "$gs\_graph$." This $gs\_graph$ serves as a straightforward intermediate representation characterized by interconnected Nodes, each functioning as an operator. Every Node maintains its own set of inputs and outputs. To enhance subsequent analysis, we construct our Computational Graph using $gs\_graph$ which is used in \cref{sec:method} as a foundation for \oursacro{}. Instead of relying solely on operator Nodes, we introduce separate nodes for operators, parameters, and intermediate data. This approach allows us to define propagation methods on the nodes we generate.

\subsection{Coupling channels via mask propagation}
\label{app:mask_propagation}
 Our approach hinges on the development of mask propagation rules tailored to individual core ONNX operators, the rules provide information on how channels are correlated within a single ONNX operator. Once these propagation rules are established for all operators within a network structure, we gain the capability to comprehensively analyze this network. By formulating rules for the majority of foundational operators, our methodology can effectively analyze a broad spectrum of neural network architectures. Furthermore, in the event that new operators are introduced, we can seamlessly extend our analysis by defining specific rules for these novel operators. This adaptability ensures our method remains versatile and up-to-date in addressing evolving neural network structures. 
 
 Our implementation supports more than 150 different operators, which are building blocks of deep learning architectures. As an example, we take the important example of the propagation through one and two General Matrix Multiplication (GeMM) operators defined by ONNX. First, we show a simplified definition of GeMM.\\

\bigbreak
\hspace{0.5cm}\textbf{Function:}
\begin{myitemize}[noitemsep,topsep=0pt]
    \item compute $Y=X*W+B$
\end{myitemize}

\hspace{0.5cm}\textbf{Inputs:}
\begin{myitemize}[noitemsep,topsep=0pt]
    \item $X$: input tensor with shape $(M, K)$
    \item $W$: input tensor with shape $(K, N)$
    \item $B$: optional input tensor, if not specified, the computation is done as if $B$ is a scalar $0$. The shape of $B$ should be unidirectional broadcastable to $(M, N)$.
\end{myitemize}

\hspace{0.5cm}\textbf{Outputs:}
\begin{myitemize}[noitemsep,topsep=0pt]
    \item $Y$: output tensor with shape $(M, N)$
\end{myitemize}

\textbf{Propagation through one GeMM operator:} We establish the propagation rule for the GeMM operator when every possible dimension (i.e. first dimension denoted by $0$ or second dimension denoted by $1$) of every possible involved variable (i.e. $X$, $W$, $B$, $Y$) is masked. Given the input mask of a single data node among all data nodes linked to the operator, the analysis procedure yields masks for the remaining data nodes. Detailed guidelines governing the analysis of GeMM are documented in \cref{tab:rule_gemm}. To illustrate, considering the first column \cref{tab:rule_gemm}, it implies that the removal of the first dimension in input $X$ necessitates the simultaneous removal of the first dimension in both $B$ and output $Y$.
\begin{table*}[htpb]
  \caption{Analysis rule of GeMM operator. Given an input mask covering dimension $0$ or $1$ of any variable $X$, $W$, $B$, $Y$, the analysis rule defines the dimensions which should be covered in the output masks for the other variables.}
  \label{tab:rule_gemm}
  \centering
  \begin{tabular}{c|c|c|c|c|c|c|c|c}
    \toprule
      Input mask  & X:0     & X:1 & W:0 & W:1      & B:0     & B:1      &Y:0     & Y:1 \\
      Output mask & B:0,Y:0 & W:0 & X:1 & B:1, Y:1 & X:0,Y:0 & W:1, Y:1 &X:0,W:0 & W:1,B:1\\
    \bottomrule
  \end{tabular}
\end{table*}

\begin{figure*}[h]
    \centering
    \includegraphics[width=0.6\textwidth]{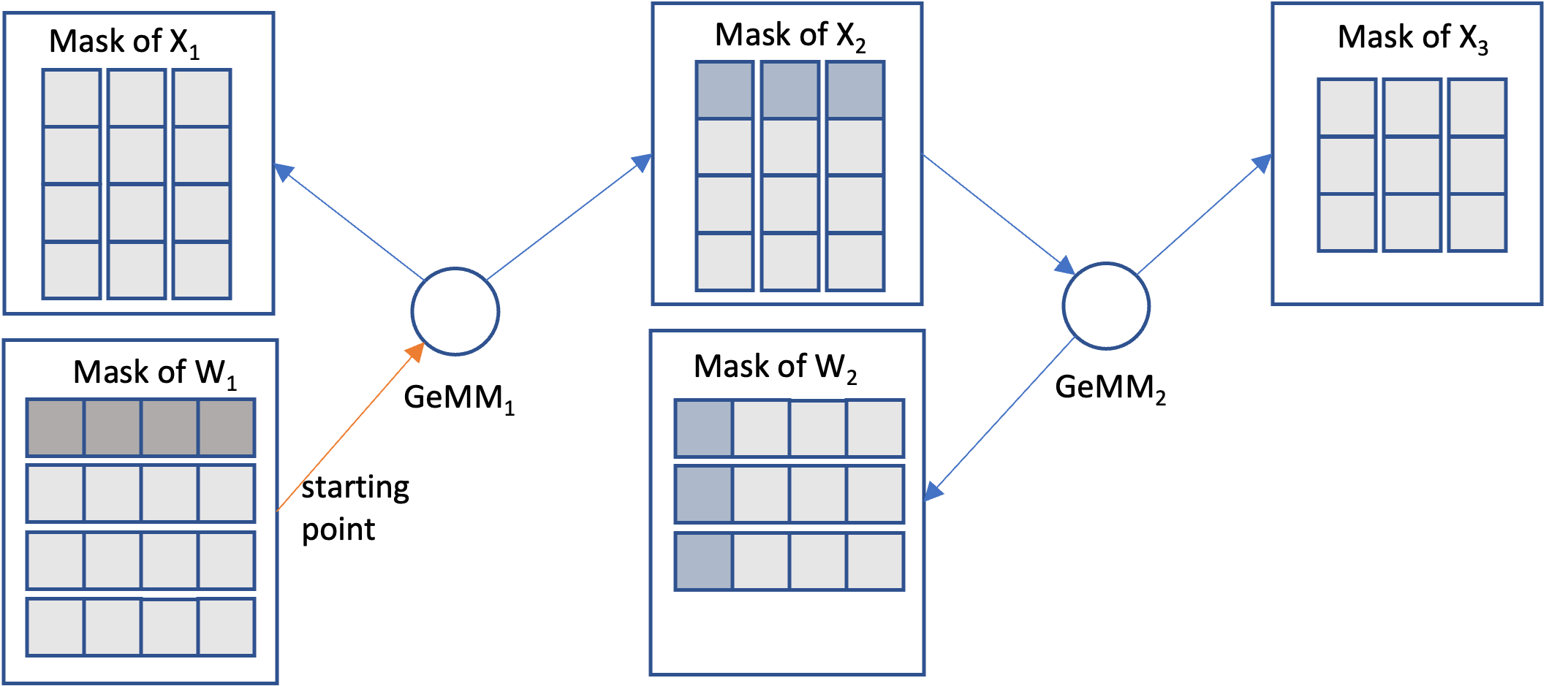}
    \caption{Example of operator-level analysis of a two connected GeMM. The analysis starts by masking the first output channels of $W_1$, through a series of mask propagation, the first feature dimension of $x_2$ and the first input channel of $W_2$ are also masked. The propagation order is illustrated through arrows.}
    \label{fig:example_mask_prop}
\end{figure*}

\textbf{Propagation through two GeMM operators:} With the analysis rule of GeMM, we then provide an illustrative depiction of our analysis applied to two connected GeMM operators in \cref{fig:example_mask_prop}. The computational graph depicts the linkage of two interconnected GeMM operators, each containing a pair of input nodes (one serving as the operator input, and the other as the weight matrix) as well as an output data node. To simplify the illustration, we consider the GeMM operator without a bias term. For input and output data nodes, each column corresponds to a distinct sample, while the row count corresponds to the number of features. For the weight nodes the column number indicates the input feature number, and the row number indicates the output feature number. As an example, $X_1$ serves as the input for GeMM$_1$, encompassing 3 samples, each possessing 4 features. The output of GeMM$_1$ comprises 4 features, hence the weights of GeMM$_1$ form a $4 \times 4$ matrix, and the resulting output, $X_2$, assumes a shape of $4 \times 3$.

The mask propagration analysis starts by applying a mask to one target channel of the source node. In \cref{fig:example_mask_prop}, we aim to eliminate the first output channel of $W_1$. The algorithm first finds the corresponding operators of this data node. In this case, GeMM$_1$ is the only operator that needs to be analyzed since $W_1$ belongs to it and there are no other operators that generate $W_1$. By applying predefined rules defined in \cref{tab:rule_gemm}, we are given a new mask of $X_2$, indicating the necessity of deleting the first feature of $X_2$, as well as the fact that $X_1$ is not affected. Then we apply the same methods on the new mask of $X_2$, it will first find both GeMM$_1$ and GeMM$_2$ as affected operators, but we will skip GeMM$_1$ since it is already analyzed. Through this analysis step, we are returned the new mask of $W_2$, which indicates that we also need to delete the first input channels of $W_2$. We are also informed that $X_3$, the output of GeMM$_2$ will not be affected. The analysis will end here because the mask of $W_2$ will incur no analysis on new operators. In this way, we get the coupled channels of the initial target channel in the form of masks. 

\subsection{Importance Estimation}
\cref{alg:group-level importance estimation} is used to assign each coupled channel an importance score.  The assessment of importance scores for individual parameters is first undertaken through designated criteria. Subsequently, the aggregation of these scores within each prunable dimension yields a consolidated measure. In pursuit of a global pruning strategy, scores are normalized within each group, thereby ensuring uniformity across all groups.

\begin{algorithm}
\caption{Group-level importance estimation}\label{alg:group-level importance estimation}
 \hspace*{\algorithmicindent} \textbf{Input} Groups: G, importance estimation criterion\\
 \hspace*{\algorithmicindent} \textbf{Output} score for each coupled channel 
\begin{algorithmic}[1]
\State assign each parameter a score with the salience estimation criterion
\State $scores = \emptyset$ \Comment{initialize score}
\For{$g$ in $G$}
        \State $scores_g = \emptyset$ \Comment{initialize score of the group}
        \For{$CC$ in $g$}
            \State $score_{CC}$ = $AGG(S(\theta_k))$ for all $\theta_k$ in $CC$
            \State $scores_g.insert(score_{CC}$)
            \EndFor
        \State $scores.insert(Norm(scores_g)$)
        \EndFor
\State return $scores$
\end{algorithmic}
\end{algorithm}

\subsection{Pruning Criteria}
Pruning requires selectively removing redundant parameters (or connections) in the neural network. In order to do so, one has to come up with a good criterion to identify such redundant connections. In this section, we introduce some popular criteria that are applied to our method.

We first introduce important notations. Assume we have a neural network $F: y = f_\Theta(x)$ with parameter $\Theta$, that maps the input data $x\in\mathbb{R}^{m}$ to the output $y \in \mathbb{R}^{n}$, $\Theta$ denotes the parameters of the neural networks, a specific parameter is denoted as $\theta$. The neural networks have multiple layers, we use $L$ to denote the total layer number and $l$ to denote a specific layer. The parameters are optimized based on the loss function $\mathcal{L}$. We use $g$ and $H$ to denote the first-order derivative and second-order derivative (Hessian) of the loss with respect to the parameters, For a specific parameter,  $g(\theta) = \frac{\partial\mathcal{L}}{\partial\theta}$, $H(\theta) = \frac{\partial^2\mathcal{L}}{\partial\theta^2}$, the importance score is $S(\theta)$.

\textbf{Magnitude-based criterion} directly uses the magnitude of each parameter as its importance score, parameters below a certain threshold are regarded as redundant. It can be simply defined as \cref{equ:criterion_L1}. 
\begin{equation}
\label{equ:criterion_L1}
    S(\theta) = |\theta|
\end{equation}

\textbf{SNIP} \cite{lee2018snip} is a sensitivity-based unstructured pruning criterion to be applied before training. To calculate the sensitivity of each parameter, an auxiliary gate variable $c$ over the model's parameter is defined. They then initialize all $c = 1$ and do not update them
anymore. the criterion is defined as the derivative of the loss w.r.t. the gates according to \cref{equ:criterion_SNIP}. 
\begin{equation}
\label{equ:criterion_SNIP}
    S(\theta) = \frac{\partial\mathcal{L}(\theta \odot c)}{\partial c} = g(\theta)\odot\theta
\end{equation}
\textbf{SNAP} \cite{verdenius2020pruning} proposed method to extend SNIP to structured pruning criterion by applying the auxiliary gates $c = 1$ over each node’s activation, which is denoted as $h$, the $i$th activation in layer $l$ is denoted as $h_i^{(l)}$ the importance score will be calculated with respect to the activation instead of a single parameter as defined in \cref{equ:criterion_SNAP}.

\begin{equation}
\label{equ:criterion_SNAP}
    S(h_i^{(l)}) = \frac{\partial\mathcal{L}(h_i^{(l)} \odot c_i^{(l)})}{\partial c_i^{(l)}}
\end{equation}

\textbf{GraSP} \cite{Wang2020Picking} is based on the second-order derivative (Hessian) of the loss with w.r.t. the parameters. The goal of GraSP is to preserve or even increase the gradient flow. The \cref{equ:criterion_GraSP} is used to measure the change of the gradient flow after pruning the parameter. If the score is positive, removing the corresponding parameter will reduce the gradient flow, and if the score is negative, removing the parameter will increase the gradient flow. 

\begin{equation}
\label{equ:criterion_GraSP}
    S(\Theta) = -\Theta^TH(\Theta)g(\Theta)
\end{equation}

\textbf{CroP} \cite{lubana2021a,earlycrop} also apples the second-order derivative to calculate the importance. The score of CroP is calculated as \cref{equ:criterion_Crop}. The idea of this criterion is to preserve the gradient flow or training dynamics during training. 
\begin{equation}
\label{equ:criterion_Crop}
    S(\Theta) = |\Theta^TH(\Theta)g(\Theta)|
\end{equation}

\textbf{Structured-GraSP} (\cref{equ:criterion_structuredGraSp}) and \textbf{Structured-CroP} (\cref{equ:criterion_structuredCrop})  apply a similar idea as SNAP to add auxiliary gate variables over activation to extend the unstructured criterion to a structured one. 

\begin{equation}
\label{equ:criterion_structuredGraSp}
    S(h^{(l)}) = -H(c^{(l)})g(c^{(l)})
\end{equation}

\begin{equation}
\label{equ:criterion_structuredCrop}
    S(h^{(l)}) = |H(c^{(l)})g(c^{(l)})|
\end{equation}

\textbf{OBD} \cite{LeCun1989OptimalBD} and \textbf{OBS} \cite{Hassibi1992SecondOD}  use the Hessian of the loss w.r.t. the parameters to calculate the importance score, the higher the value of Hessian, the higher the importance of the parameters. For the $j$th parameter $\theta_j$, see \cref{equ:criterion_OBD} for OBD score and \cref{equ:criterion_OBS} for OBS score. However, this approach requires the calculation of Hessian of all parameters of the neural networks, making it intractable to compute for large networks.

\begin{equation}
\label{equ:criterion_OBD}
    (OBD)\qquad S(\theta_j) = \frac{\theta_j^2H_{j,j}}{2}
\end{equation}

\begin{equation}
\label{equ:criterion_OBS}
    (OBS)\qquad S(\theta_j) = \frac{\theta_j^2}{2H_{j,j}^{-1}}
\end{equation}

\textbf{OBC} \cite{frantar2022obc} applies the method of OBS layer-wise to make the calculation tractable. Instead of minimizing the influence on the final loss in OBS, OBC minimizes the reconstruction error per layer, see \cref{equ:layerwise} for problem definition, the goal is to find the optimal weight mask as well as an optimal update of the weight matrix to minimize the reconstruction error. OBC introduces a greedy solver that removes weights one-at-a-time, then fully reconstruct the remaining weights after each iteration via an efficient closed-form equations. The importance of the $j$th parameter of the $l$th layer is determined by their influence on the reconstruction error of the layer output as defined in \cref{equ:criterion_layer-OBS}. The hessian matrix of each layer is used here to calculate the importance and to update the parameters after pruning, they can be derived by taking the outer product of the calibration data per layer as $H^{(l)} = X^{(l)}X^{(l)T}$.

\begin{equation}
\label{equ:criterion_layer-OBS}
    S(\theta_j^{(l)}) = \frac{(\theta_j^{(l)})^2}{[(H^{(l)})^{-1}]_{j,j}}
\end{equation}

\subsection{OB\oursacro{} and SparseGPT}

\begin{figure*}[!h]
    \centering
    \begin{subfigure}[b]{0.8\textwidth}
        \centering
        \includegraphics[width=\textwidth]{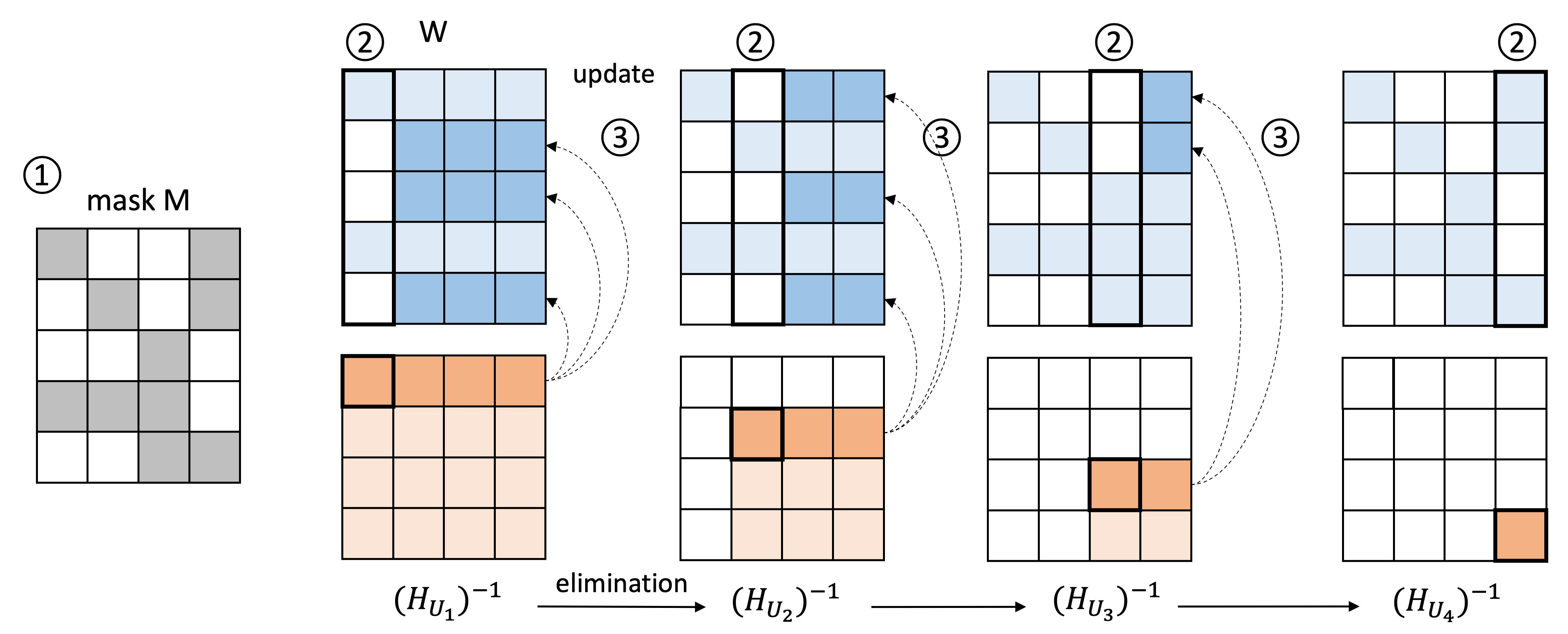}
        \caption{SparseGPT}
        \label{fig:SparseGPT}
    \end{subfigure}
    \begin{subfigure}[b]{0.8\textwidth}
        \centering
        \includegraphics[width=\textwidth]{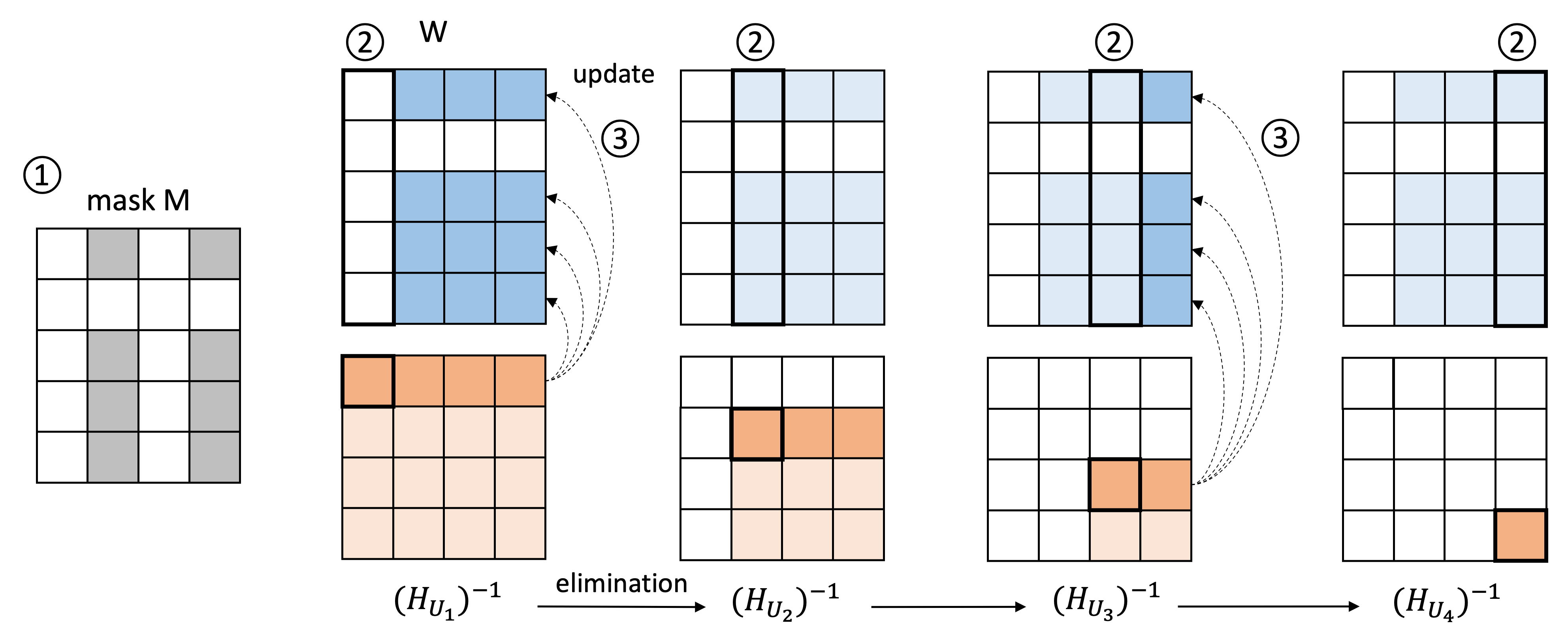}
        \caption{OBSPA}
        \label{fig:SPA-OBC}
    \end{subfigure}
    \caption{Visualization of reconstruction algorithm of \cite{Frantar2023SparseGPTML} and OB\oursacro{}. \textcircled{1} mask are derived according to layer-OBS score. for SparseGPT, zeros are scattered in the mask while for OB\oursacro{}, zeros span the whole channel. \textcircled{2} weights in the first column of the weight matrix are pruned. \textcircled{3} Using Hessian inverses $(H_{u_j})^{-1}$ to update the reminder of the weight (only in dark blue). Then repeat \textcircled{2}\&\textcircled{3} for the next column until all columns are processed}
    \label{fig:compare_SparseGPT_SPA-OBC}
\end{figure*}

\textbf{SparseGPT} \cite{Frantar2023SparseGPTML} is a large-scale extension of OBC that proposes a method to incrementally prune weights in each column of the weight matrix. Different from OBS that uses the whole Hessian of the layer to adjust the values of all available parameters to compensate for the removal, \cite{Frantar2023SparseGPTML} only updates the weight among the remaining unpruned weights with a smaller Hessian matrix. The update procedure of SparseGPT is illustrated in \cref{fig:SparseGPT}. 

\textbf{OB\oursacro{}} is extended to a structured pruning algorithm from SparseGPT by applying group-level importance estimation and directly masking entire columns and rows before structurally deleting them. In OB\oursacro{}, we determine the coupled channels to be pruned by applying the layer-OBS \cite{frantar2022obc} criterion and then create masks for those channels. We then apply the masks on the weight matrix column by column and update the remaining columns. For a specific column $i$ that needs to be pruned, we first calculate the error and then update the remaining parameters with the following equations. 
\begin{equation}
    err = \frac{\Theta_{:,i}}{H^{-1}_{i,i}}
\end{equation}
\begin{equation}
    \Theta_{:,i:} = \Theta_{:,i:}-err\cdot H^{-1}_{i,i:} 
\end{equation}

\subsection{Implementation details}
\begin{figure*}
    \centering
    \includegraphics[width=\textwidth]{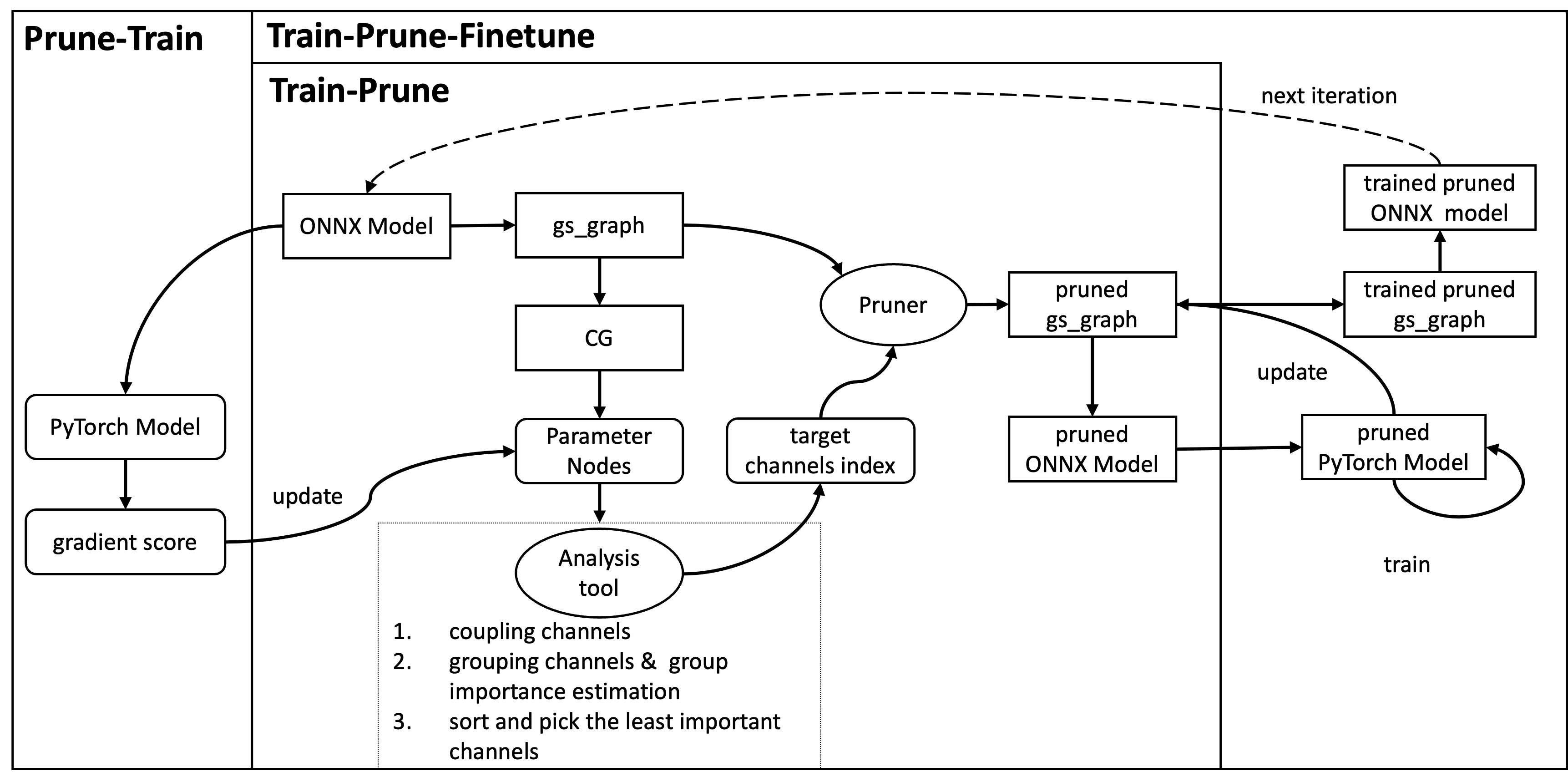}
    \caption{Detailed implementation of \oursacro{}}
    \label{fig:implementation}
\end{figure*}

We provide \cref{fig:implementation} for a compact overview of the implementation of our method. As detailed in previous sections, we first obtain a $gs\_graph$ and build our Computational Graph based on it. Then we apply mask propagation and importance estimation on the computational graph to derive the index of target channels for pruning.  We can then very conveniently prune those channels on $gs\_graph$ and convert $gs\_graph$ to ONNX model using tools provided by onnx-graphsurgeon. In this way, we can already support the Train-Prune framework. To further support Train-Prune-Fintune and Prune-Train settings, we add additional blocks to convert ONNX model to PyTorch model. This conversion grants our method the ability to apply sensitivity-based criteria and to train/fine-tune the pruned model. 

\section{Experiments Details}

\subsection{Dataset Details}

\textbf{CIFAR-10} \cite{CIFAR10} and \textbf{CIFAR-100} \cite{CIFAR100} datasets both serve as platforms for image classification tasks, diverging based on their class count and intricacy. CIFAR-10 comprises a collection of 60,000 32x32 color images, categorized into ten distinct classes, each containing 6,000 images. These classes encompass common objects like airplanes, automobiles, birds, cats, dogs, and more. In contrast, CIFAR-100, also consisting of 60,000 images, exhibits a finer granularity with 100 distinct classes, representing more nuanced categories. Notably, CIFAR-10 and CIFAR-100 are mutually exclusive, allowing for a reciprocal utilization wherein CIFAR-100 serves as an out-of-distribution dataset for CIFAR-10, and vice versa.

\textbf{ImageNet-1k} \cite{deng2009imagenet}, is a widely recognized and extensively used dataset in the field of computer vision and machine learning. This dataset consists of millions of labeled images, each categorized into one of the 1,000 predefined classes or object categories. The diversity and size of ImageNet make it a valuable resource for training and evaluating deep learning models. We also evaluate OB\oursacro{}'s performance on ImageNet-1k. 

\textbf{ImageNet-O} \cite{hendrycks2021naeimagnet-o} dataset is the natural adversarial example dataset for out-of-distribution detectors of ImageNet-1k. It consists of 2000 images from 200 classes that are not found in the ImageNet-1k dataset. We resize the images to 224x224. This dataset serves as an OOD dataset of ImageNet-1k in our experiment.

\textbf{Imagenette} \cite{imagewang}, derived from ImageNet, showcases 13394 images from a subset of 10 easily classifiable classes (e.g., tench, English springer, cassette player, chain saw, church, French horn, garbage truck, gas pump, golf ball, parachute). We also preprocess the images to 224x224. Despite its modest size, Imagenette proves to be a suitable testbed to assess the functionality of \oursacro{} across models with divergent architectures.

\textbf{SST-2} \cite{SocherEtAl2013:RNTN}, the Stanford Sentiment Treebank 2 (SST-2) is a popular dataset for sentiment analysis in natural language processing. It consists of 215,154 unique phrases from movie reviews, where each review is labeled with its sentiment as either "positive" or "negative". The dataset is well-structured, and it has been widely used for training and evaluating sentiment analysis models. In our work, we use pruning a DistilBERT model on this dataset to show \oursacro{}'s ability to prune self-attention-based NLP models.

\subsection{Metric Details}
Reduction in Floating Points Operations and Reduction in Parameters are widely used in many papers \cite{narshana2023dfpc,fang2023depgraph} to demonstrate the effectiveness of pruning methods, we provide definition of these two evaluation metrics
\begin{enumerate}
    \item Reduction in Floating Point Operations, represented as RF, quantifies the acceleration in FLOP execution speed achieved through pruning.
    \begin{equation}
        RF = \frac{FLOP_{before}}{FLOP_{after}}
    \end{equation}
    \item Reduction in Parameters, denoted as RP, evaluates the parameter reduction achieved through the pruning process.
    \begin{equation}
        RP = \frac{\#params_{before}}{\#params_{after}}
    \end{equation}
\end{enumerate}

\subsection{Setting Details}
\label{sec:setting_details}

For the experiment that follows the Train-Prune-Finetune and Prune-Train schemes on CIFAR and ImageNette datasets, we use a 12GB NVIDIA GeForce GTX 1080 Ti GPU, for the experiments that follow the Train-Prune setting and the experiment on ImageNet, we use a 40G NVIDIA A100 GPU.

\textbf{Prune any framework:} We test the framework-agnostic ability of \oursacro{} on the ImageNet dataset. We first define random initialized ResNet-18 from PyTorch, TensorFlow, JAX, and MXNet respectively, they are then trained for 100 epochs on their original frameworks before being converted to ONNX. While PyTorch, TensorFlow, and MXNet offer direct conversion functionalities, Jax models necessitate an additional intermediary step, involving a conversion to TensorFlow before arriving at the ONNX representation. Then we prune and finetune the model based on \oursacro{}-L1. In addition to the pruning outcomes, we also test the computational overhead incurred during the framework conversion process. We quantify this overhead by reporting the average model conversion time, derived from 10 separate conversion instances as shown in \cref{tab:model_conversion_time}.

\textbf{Prune any architecture:} The functional test of architecture-agnostic property of \oursacro{} is done on both CFIAR10 and SST-2. Here we conducted pruning experiments on DenseNet-121 \cite{Huang2016DenselyCC}, EfficientNet-b0 \cite{pmlr-v97-tan19a} MobileNet-v2 \cite{Howard2017MobileNetsEC}, RegNet\_x\_16gf \cite{Radosavovic2020DesigningND}, ResNet-18 \cite{He2015DeepRL}, Resnext-50\_32x4d \cite{Xie2016AggregatedRT}, VGG-16 \cite{Simonyan15}, Wide-ResNet-101\_2 \cite{Zagoruyko2016WRN} sourced from TorchVision, VIT \cite{dosovitskiy2021an} and DistilBERT \cite{devlin-etal-2019-bert} sorced from HuggingFace. The setting of those experiments are same as framework-agnotic experiments. 

\textbf{Prune with fine-tuning:} This set of experiments is first done on ResNet-18 and VGG-16 to perform image classification on CIFAR-10 and CIFAR-100. We compare the L1-based method \oursacro{}-L1 to its ungrouped counterpart for the Train-Prune-Finetune setting and compare \oursacro{}-SNIP, \oursacro{}-CroP and \oursacro{}-GraSP to their structured algorithms, SNAP, Structured-CroP and Structured-GraSP for Prune-Train setting. To ensure equitable comparisons, we maintain uniformity in total epochs across all configurations. When pruning is executed after training, the model undergoes 100 epochs of training followed by 100 epochs of pruning and fine-tuning. Conversely, for pruning before training, a total of 200 epochs is allocated for the combined pruning and fine-tuning procedure. Besides, building upon the findings in \cite{verdenius2020pruning}, which advocate for the efficacy of iterative pruning, we conduct iterative experiments for each criterion. In this iterative version, we employ 5 steps, with 5 training epochs between each step. The optimization procedure involves the use of the SGD optimizer and CosineAnnealingLR as the learning rate scheduler.

For the experiments on ImageNet-1k, we first pruned pre-trained ResNet-50, DenseNet-121, and Vit\_b\_16 using \oursacro{}-L1 and OB\oursacro{}, we then fine-tune the models following \cite{fang2023depgraph}'s setting, with 90 epochs of fine-tuning on both pruned ResNet-50 and DenseNet-121. However different from \cite{fang2023depgraph} that fine-tunes ViT for 300 epochs, we only fine-tune ViT for 30 epochs. We also follow \cite{leclerc2023ffcv} to perform fast training. 

\textbf{Prune without fine-tuning:} We follow the setting from DFPC to evaluate the performance of ResNet-50, ResNet-101, and VGG-19 models on both CIFAR-10 and CIFAR-100 datasets under the pruning without fine-tuning scheme. Models are pre-trained before pruning and no further fine-tuning is allowed after pruning. We also conduct experiments on the ImageNet-1k dataset. We use calibration data to calculate the Hessian per layer for importance estimation and parameter update. for the CIFAR dataset in which samples are in low resolution, 2048 data samples are used, for the ImageNet dataset, 896 data points are used.  For image classification tasks, \oursacro{}-OBC encompasses two distinct settings: data-driven setting and data-free setting. Data points are directly sampled from the training set in the data-driven setting, but in the data-free setting, calibration data are either drawn from the OOD dataset or generated following a uniform distribution between 0 and 1. CIFAR-10 and CIFAR-100 are mutually exclusive, they can serve as OOD datasets for each other, we also use ImageNet-O as the OOD dataset of ImageNet-1k. However, In NLP tasks, where different sentences can be easily accessed, choosing random sentences is not rational. Consequently, we exclusively utilize out-of-distribution (OOD) datasets. Specifically, we employ the ax dataset as an example of an OOD dataset of SST-2.

We also need to mention an additional noteworthy observation pertaining to the performance enhancement achieved through the resetting of batch normalization statistics following pruning, a phenomenon previously elucidated in OBC \cite{frantar2022obc}. In our study, we adopt a straightforward approach of forwarding the calibration data twice to facilitate the updating of running mean and running variance in the batch normalization layers. However, it is important to highlight that this performance gain is exclusively relevant to the ID and OOD settings. The presence of informative calibration data in these scenarios enables effective updates of batch normalization statistics. In contrast, when employing randomly generated calibration data, the batch normalization statistics can become distorted, leading to potential performance degradation. Therefore, in this experimental
context, we implement batch normalization statistic re-calibration exclusively for the ID and OOD scenarios, while refraining from its utilization in the data-free setting.

\section{Additional Experiment}
\subsection{Model Framework Conversion Time}
We test the conversion time from different frameworks to ONNX. The results, as detailed in \cref{tab:model_conversion_time}, reveal that even for the Jax models requiring dual conversions, the process completes within seconds. This indicates that the computational overhead incurred during the model conversion process is trivial compared to the time in pruning and training. 
\begin{table}[htpb]
  \caption{Model Conversion time from different frameworks (Pytorch, TensorFlow, MXNet, Jax) to ONNX. }\label{tab:model_conversion_time}
  \centering
  \begin{tabular}{l l l l l}
    \toprule
      &   \multicolumn{3}{c}{conversion time (s)}   \\
    Model   & PyTorch & TensorFlow & MXNet & Jax \\
    \midrule
      ResNet-18 & 0.51s & 2.47s & 2.28s & 5.47s \\
      ResNet-50 & 2.01s& 7.35s & 7.36s & 12.52s\\
    \bottomrule
  \end{tabular}
\end{table}

\subsection{\oursacro{} with fine-tuning}

\begin{figure*}
\vspace{-0mm}
    \centering
    \begin{subfigure}[b]{0.21\textwidth}
        \centering
        \includegraphics[width=\textwidth]{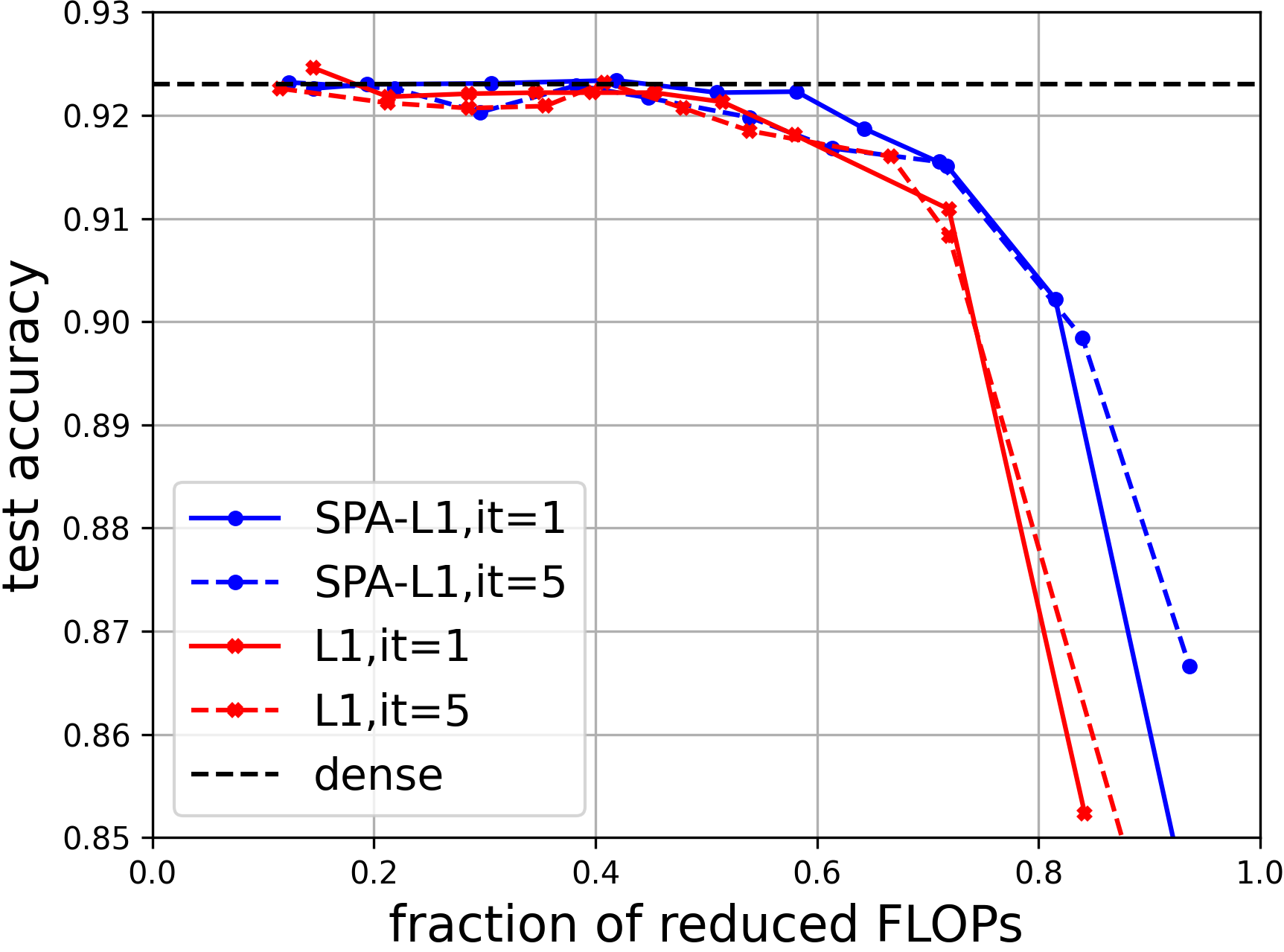}
        \caption{}
        \label{fig:res18_L1_RF}
    \end{subfigure}
    \hfill
    \begin{subfigure}[b]{0.21\textwidth}
        \centering
        \includegraphics[width=\textwidth]{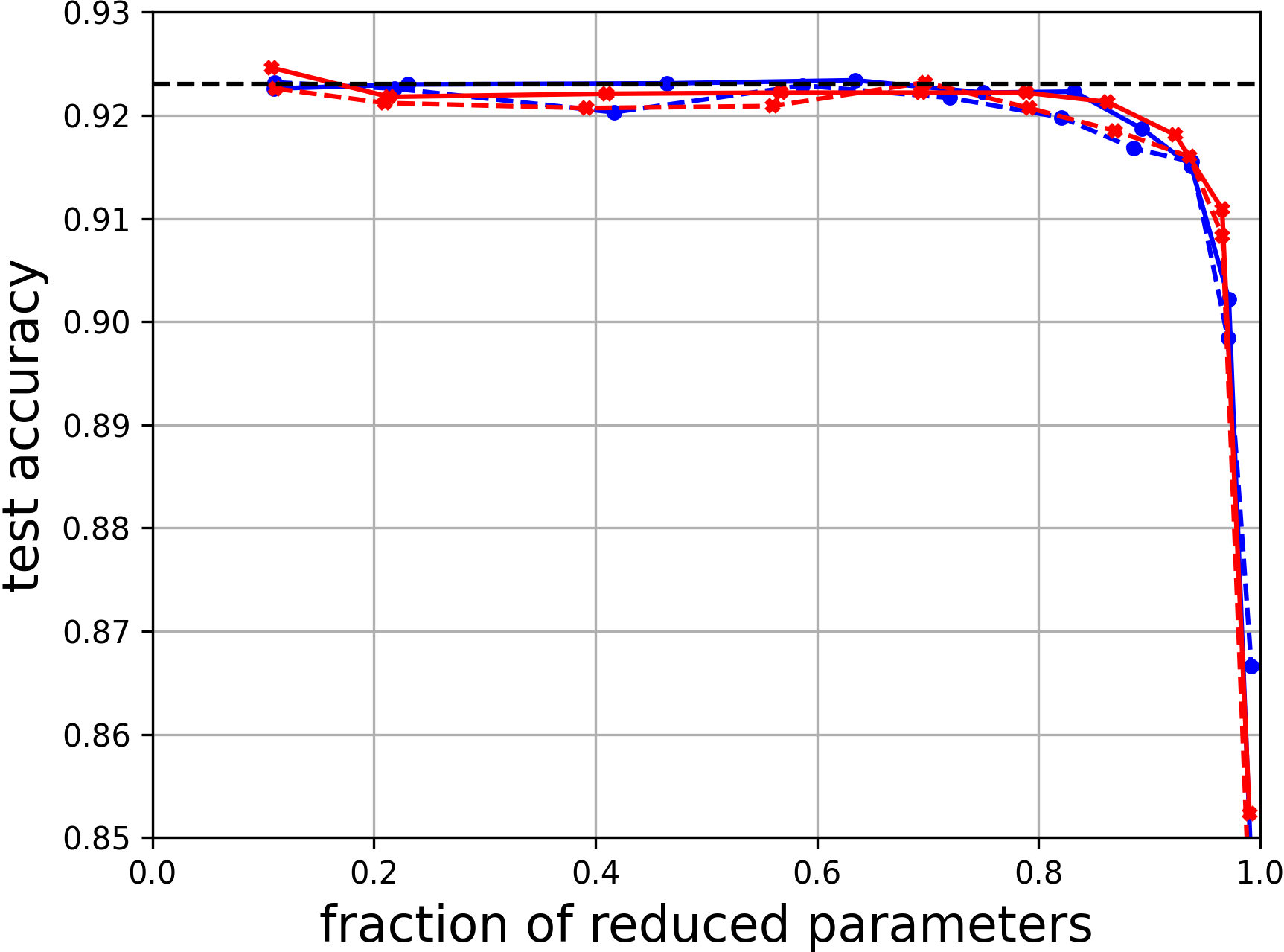}
        \caption{}
        \label{fig:res18_L1_RP}
    \end{subfigure}
    \hfill
    \begin{subfigure}[b]{0.21\textwidth}
        \centering
        \includegraphics[width=\textwidth]{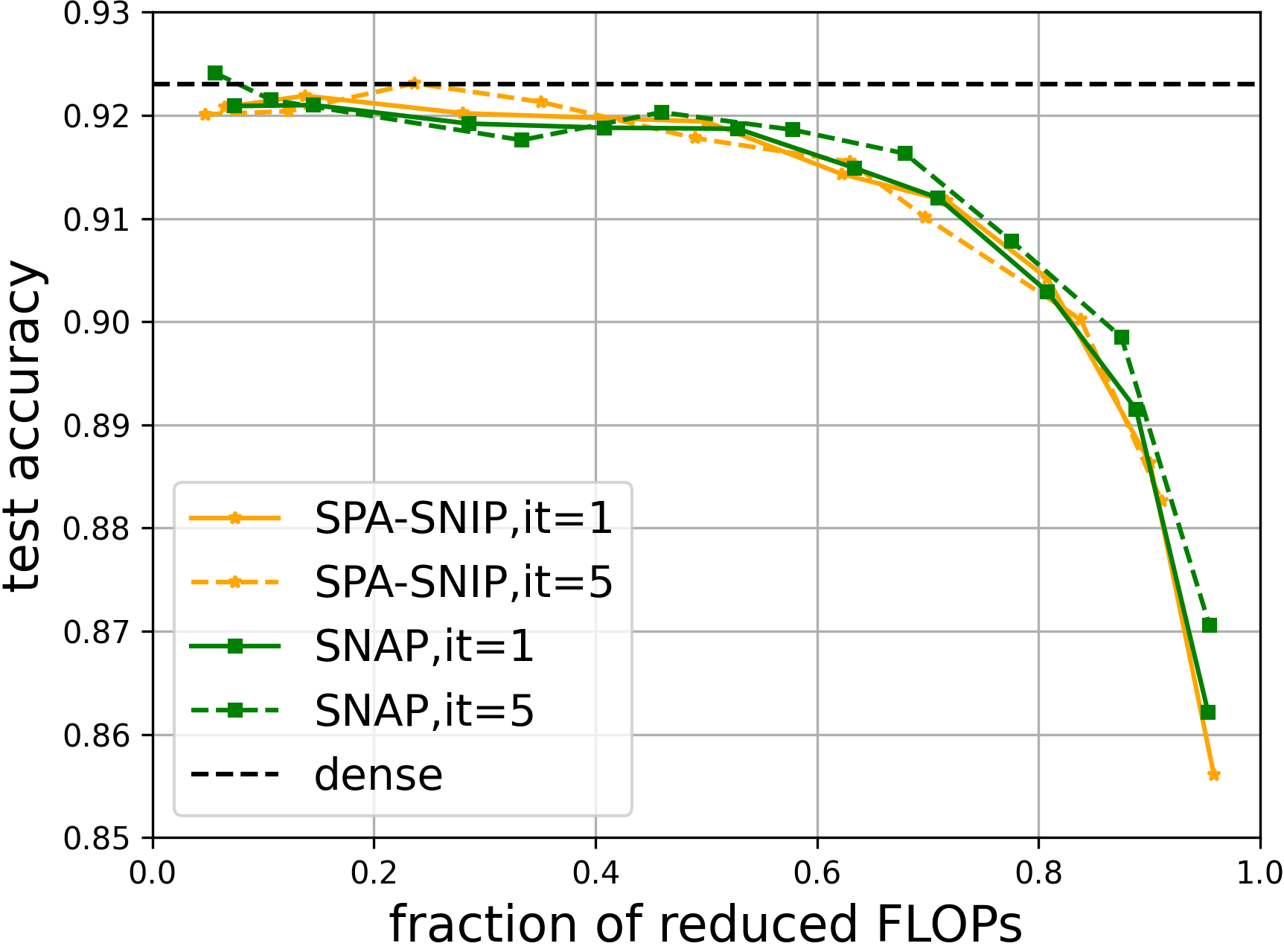}
        \caption{}
        \label{fig:res18_SNIP_RF}
    \end{subfigure}
    \hfill
    \begin{subfigure}[b]{0.21\textwidth}
        \centering
        \includegraphics[width=\textwidth]{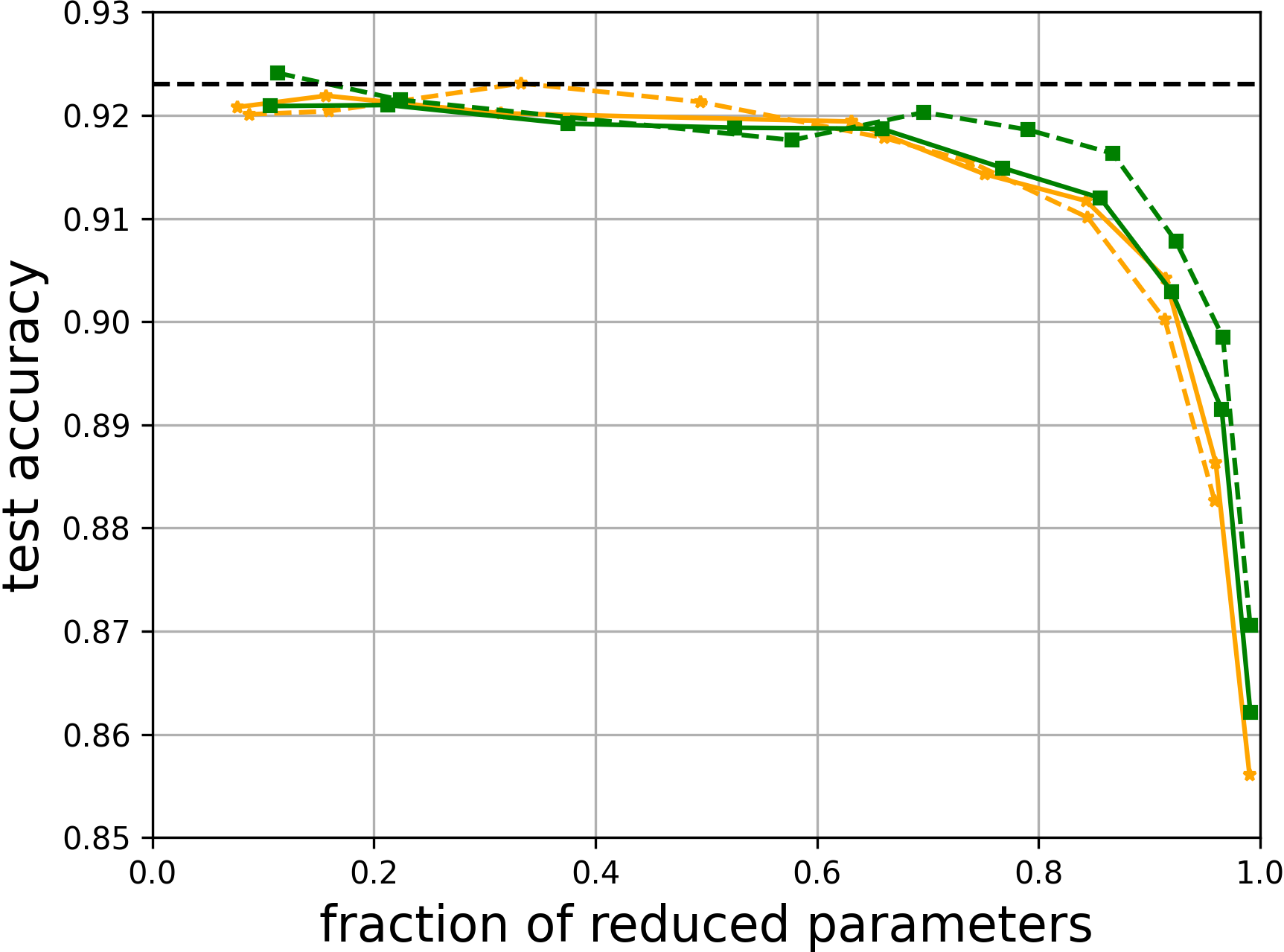}
        \caption{}
        \label{fig:res18_SNIP_RP}
    \end{subfigure}
    \newline
    \begin{subfigure}[b]{0.21\textwidth}
        \centering
        \includegraphics[width=\textwidth]{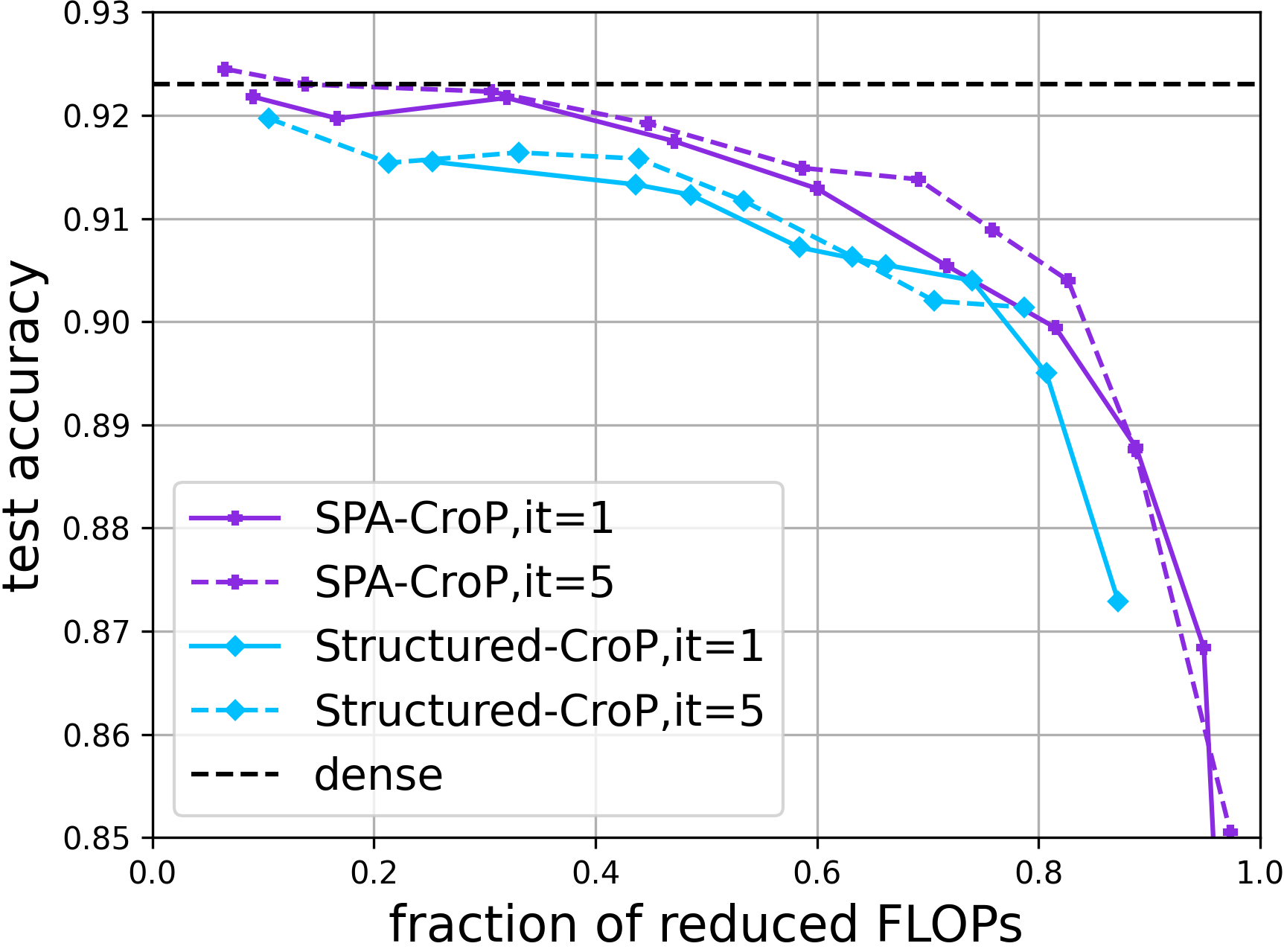}
        \caption{}
        \label{fig:res18_CROP_RF}
    \end{subfigure}
    \hfill
    \begin{subfigure}[b]{0.21\textwidth}
        \centering
        \includegraphics[width=\textwidth]{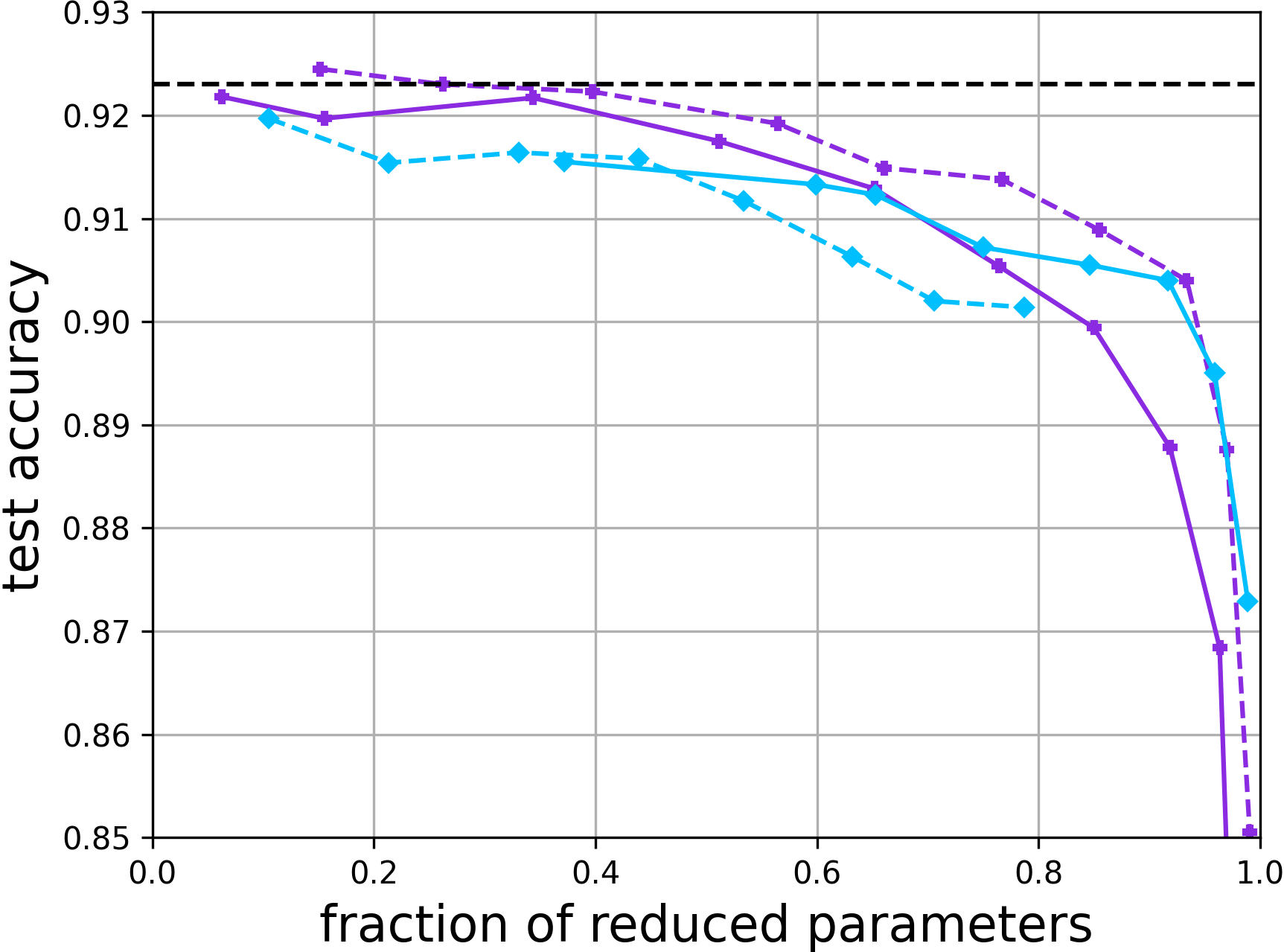}
        \caption{}
        \label{fig:res18_CROP_RP}
    \end{subfigure}
    \hfill
    \begin{subfigure}[b]{0.21\textwidth}
        \centering
        \includegraphics[width=\textwidth]{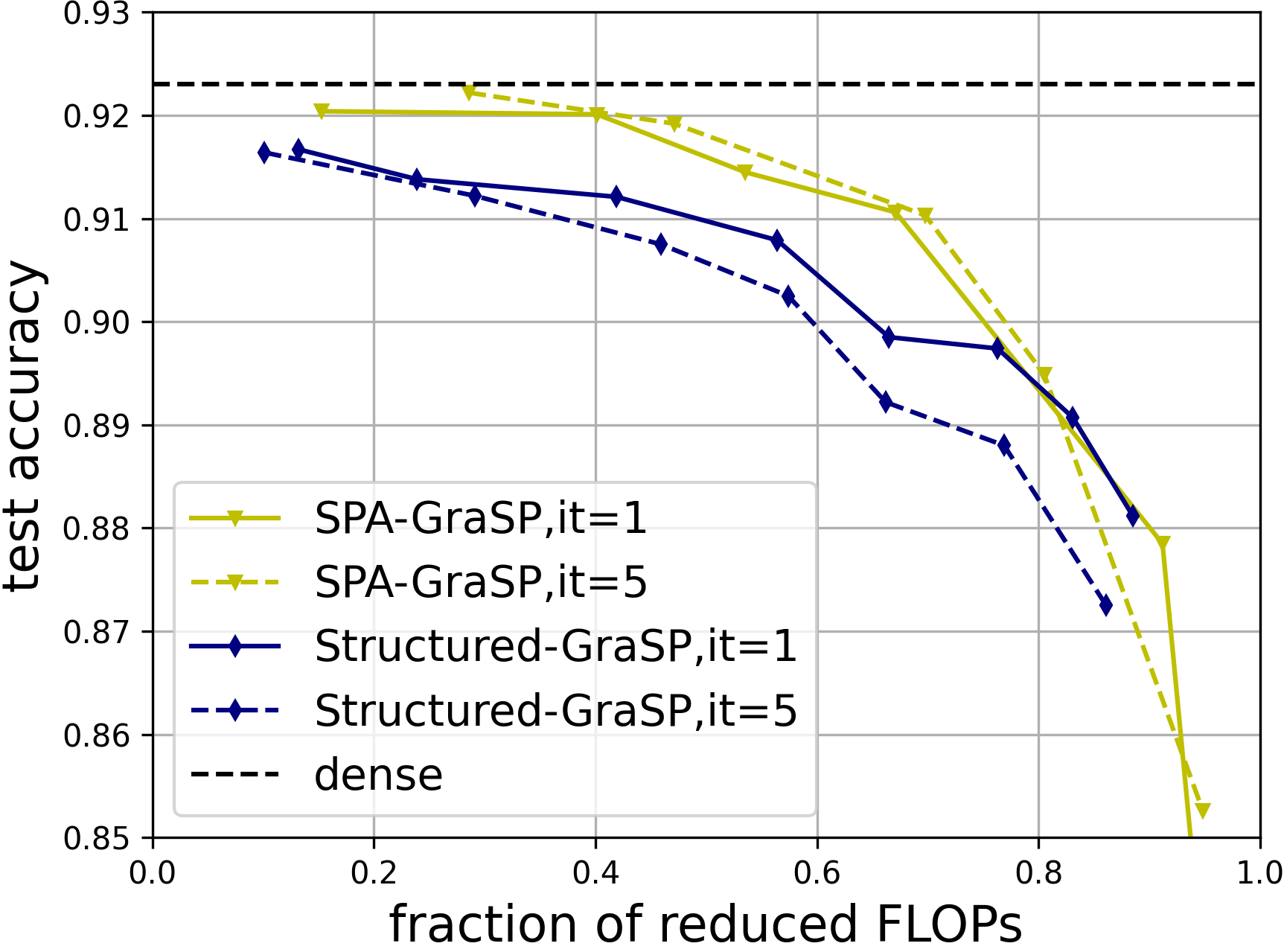}
        \caption{}
        \label{fig:res18_GRASP_RF}
    \end{subfigure}
    \hfill
    \begin{subfigure}[b]{0.21\textwidth}
        \centering
        \includegraphics[width=\textwidth]{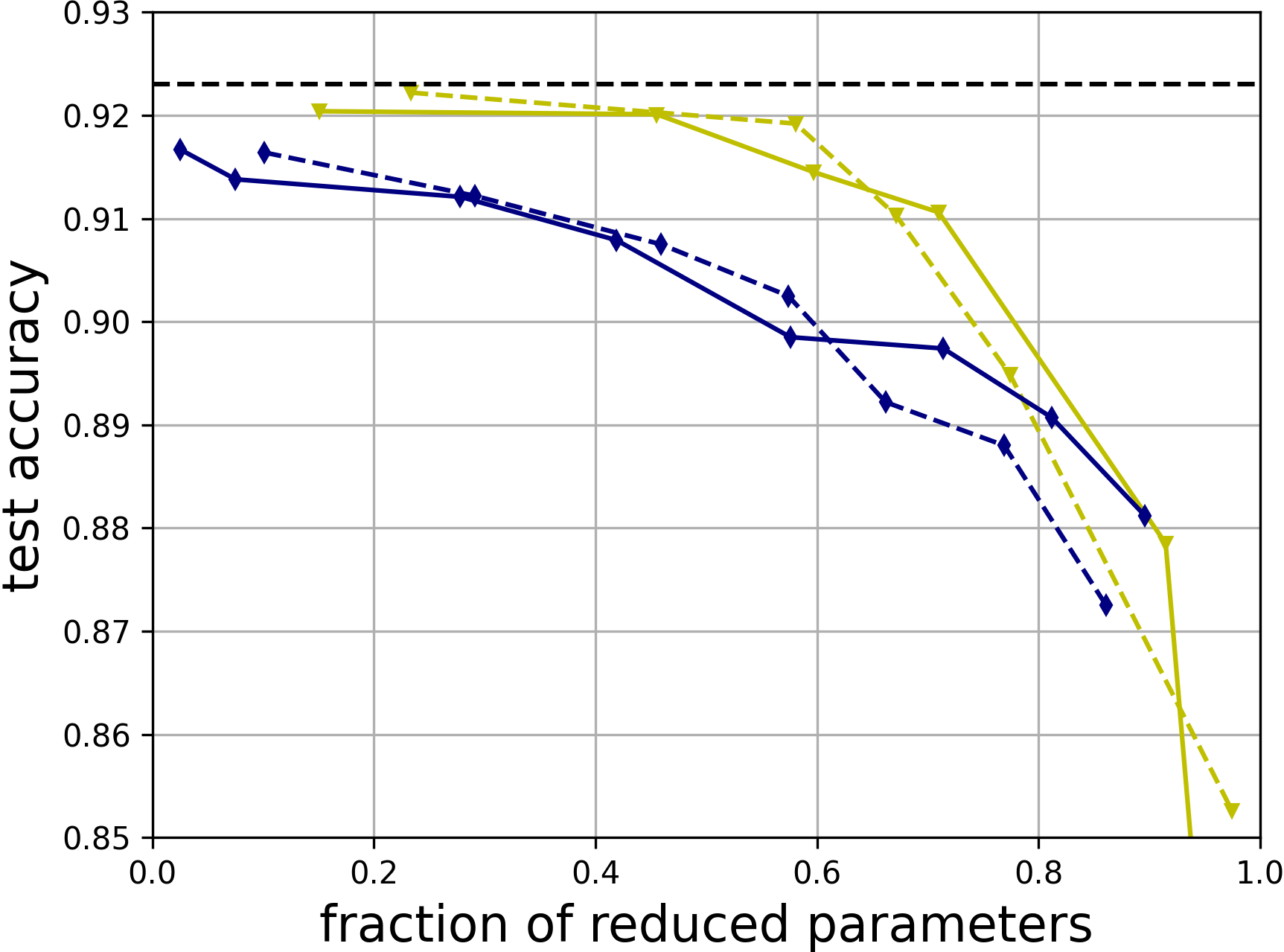}
        \caption{}
        \label{fig:res18_GRASP_RP}
    \end{subfigure}
    \vspace{-2mm}
    \caption{Trade off between accuracy and FLOPs/parameters with ResNet-18 on CIFAR-10 (see \cref{fig:res18_L1_RF,fig:res18_L1_RP,fig:res18_SNIP_RF,fig:res18_SNIP_RP,fig:res18_CROP_RF,fig:res18_CROP_RP,fig:res18_GRASP_RF,fig:res18_GRASP_RP}). \oursacro{} efficiently implements both the structured and grouped versions of train-prune-finetune criteria like L1 and prune-train criteria like SNAP, CroP and GraSP}
    \label{fig:prune-any-time-cifar10}
    \vspace{-1mm}
\end{figure*}

In this section, we report the additional exeriment results of performing pruning with \oursacro{}. \cref{fig:prune-any-time-cifar10} compares the \oursacro{} grouped versions of L1, SNIP, CroP, and GraSP to their original structured counterparts of L1, SNAP, Structured-CroP, Structured- GrasP on ResNet18 on CIFAR18. We observed that \oursacro{} versions of these pruning cirteria always matches our outperforms their strucutred versions. Further, \cref{tab:structured-densenet121-imagenet} and \cref{tab:structured-vit-imagenet} shows additional results on DenseNet/ImageNet, Vit/ImageNet. Note the in the Vit Experiment, we only fine-tuned 30 epochs after pruning while DepGraph fine-tuned 300 epochs. We observe that \oursacro{} matches or outperform other previous methods.

\begin{table}[htpb]
  \caption{Structured pruning of DenseNet-121 on ImageNet with fine-tuning. "N/R" indicate non-reported results in original papers.}\label{tab:structured-densenet121-imagenet}
  \centering
  \begin{tabular}{l | c c c c }
    \toprule
     method & top1 acc. &top5 acc. &  RF & RP \\
    \midrule
      Base Model & 74.43\% & 91.97\% & 1$\times$ & 1$\times$  \\
      DepGraph \cite{fang2023depgraph} & 73.98\%& N/R & 2.09$\times$& N/R\\
      \oursacro{}-L1 & 74.39\% &92.19\% &2.09$\times$ & 1.80$\times$ \\
      OB\oursacro{} & 74.62\% & 92.19\%&1.78$\times$ & 1.84$\times$\\
    \bottomrule
  \end{tabular}
\end{table}
\begin{table}[htpb]
  \caption{Structured pruning of ViT\_b\_16 on ImageNet with fine-tuning. ”N/R” indicate non-reported results in original papers.}\label{tab:structured-vit-imagenet}
  \centering
  \begin{tabular}{l | c c c c }
    \toprule
     method & top1 acc. &top5 acc. &  RF & RP \\
    \midrule
      Base Model & 81.43\%& 96.02\%& 1$\times$ & 1$\times$  \\
      DepGraph +EMA \cite{fang2023depgraph} &79.58\% & N/R & 1.69$\times$ & N/R\\
      DepGraph \cite{fang2023depgraph} & 79.17\%& N/R & 1.69$\times$& N/R\\
      \oursacro{}-L1 & 78.81\% &94.20\% &2.03$\times$ & 2.05$\times$ \\
      OB\oursacro{} & 78.90\% & 94.30\%&1.95$\times$ & 1.98$\times$\\
    \bottomrule
  \end{tabular}
\end{table}

\subsection{\oursacro{} without fine-tuning}
\label{app:without-fine-tuning}

\begin{table}[htpb]
\small
  \caption{Structured pruning of ResNet-101 on CIFAR-10 without finetuning}
  \label{tab:structured-oneshot-res101_1}
  \centering
  \begin{tabular}{l | c c c }
    \toprule
    &   \multicolumn{3}{c}{CIFAR-10}  \\
     method & acc. drop &  RF & RP \\
     
    \midrule
      DFPC 								    & 4.95\% & 1.64x & 2.22x  \\
      OB\oursacro{} (ID)  		& 0.93\% & 1.59x & 1.49x  \\
      OB\oursacro{} (OOD)  		& 1.08\% &  1.59x & 1.49x \\
      OB\oursacro{} (DataFree)  & 1.51\% & 1.58x & 1.49x  \\
    \bottomrule
  \end{tabular}
\end{table}
\begin{table}[htpb]
\small
  \caption{Structured pruning of ResNet-101 on CIFAR-100 without finetuning}
  \label{tab:structured-oneshot-res101_2}
  \centering
  \begin{tabular}{l | c c c }
    \toprule
    &     \multicolumn{3}{c}{CIFAR-100}\\
     method & acc. drop &  RF & RP \\
     
    \midrule
      DFPC 	 &  9.40\%& 1.72x & 1.53x\\
      OB\oursacro{} (ID)  &	 7.31\%& 1.68x & 1.51x\\
      OB\oursacro{} (OOD) & 6.68\%& 1.68x& 1.51x\\
      OB\oursacro{} (DataFree) &  9.95\%& 1.61x & 1.47x\\
    \bottomrule
  \end{tabular}
\end{table}
\begin{table}[htpb]
  \caption{Accuracy of Base Models of OB\oursacro{} experiment}
  \label{tab:base models}
  \centering
  \begin{tabular}{l | c c | c c }
    \toprule
    &   \multicolumn{2}{c|}{CIFAR-10}  &  \multicolumn{2}{c}{CIFAR-100} \\
     Model & DFPC &  ours & DFPC & ours \\
       
    \midrule
      ResNet-50 &  94.99\% & 94.70\%& 78.85\%& 78.10\%\\
      ResNet-101 & 95.09\% & 94.48\%& 79.43\%&81.05\%\\
      VGG-19 &  93.50\% & 96.04\%& 72.02\%&81.05\%\\
    \bottomrule
  \end{tabular}
\end{table}

\textbf{OB\oursacro{} with ResNet-101 and Based Models.} In this section, we first report the additional experiment result of performing pruning after training with OB\oursacro{} on ResNet-101. These results are detailed in \cref{tab:structured-oneshot-res101_1} and \cref{tab:structured-oneshot-res101_2}. We then provide the test accuracy of the base models used in our OB\oursacro{} and DFPC in \cref{tab:structured-oneshot-res50-vgg19,tab:structured-oneshot-res101_1,tab:structured-oneshot-res101_2} as \cref{tab:base models}.

\textbf{OB\oursacro{} on ImageNet-1k.}
We also conduct pruning experiments without fine-tuning on the harder ImageNet-1k. DFPC does not present results for ImageNet without fine-tuning. We observed that, while using only less than 1000 calibration data samples or no calibration data, \oursacro{} presents non-trivial compression capabilities being able to maintain accuracy above 70\% accuracy. 
\begin{table}[!htpb]
  \caption{Structured pruning of ResNet-50 on ImageNet without fine-tuning}\label{tab:structured-oneshot-res50-imagenet}
  \centering
  \begin{tabular}{l | c c c }
    \toprule
     method & accuracy &  RF & RP \\
    \midrule
      Base Model & 76.15\% & 1x & 1x  \\
      OB\oursacro{} (ID) - Low compression & 74.27\% & 1.22$\times$ & 1.16$\times$ \\
      OB\oursacro{} (ID) - High compression & 70.57\% & 1.43$\times$ & 1.20$\times$\\
      OB\oursacro{} (OOD)  - Low compression & 71.60\% & 1.25$\times$ & 1.18$\times$\\
      OB\oursacro{} (DataFree) - Low compression & 70.13\% & 1.21$\times$ & 1.19$\times$\\
    \bottomrule
  \end{tabular}
\end{table}

\subsection{Pruning Time of OB\oursacro{}}
 We compare the pruning time of our OB\oursacro{} algorithm to DFPC. The total pruning time of OB\oursacro{} includes all the necessary steps including building the computational graph, analyzing groups and applying OB\oursacro{} to prune and update parameters. For pruning a ResNet-50 on CIFAR-10 or CIFAR-100, DFPC takes 12 minutes, but our algorithm only takes 1.5 to 2 minutes. Pruning larger networks such as ResNet-101 and VGG-19 could also be completed within 6 minutes. For ImageNet-1k, a higher resolution dataset, DFPC also takes $6\times$ more time than ours OB\oursacro{}. 

The calibration data is processed batch by batch, so the batch size and batch number could also influence the pruning time. In our experiment, we use 2 batches of calibration data with batch size equal to 1024 in the CIFAR experiment and 7 batches of 128 data in the ImageNet-1k experiment.
\begin{table}[htpb]
  \caption{Pruning time for OB\oursacro{} and DFPC}\label{tab:time-structured-oneshot}
  \centering
  \begin{tabular}{l c c c}
    \toprule
     Method & Model & Dataset & Pruning time \\
    \midrule
      DFPC & ResNet-50 & CIFAR-10/100  & 12 min  \\
      DFPC & ResNet-50 & ImageNet-1k & 38 min \\
      OB\oursacro{} & ResNet-50 &  CIFAR-10/100 &  1.5-2 min  \\
      OB\oursacro{} & ResNet-101 &  CIFAR-10/100 & 3-6 min   \\
      OB\oursacro{} & VGG-19 &  CIFAR-10/100 &  3.5-4.5 min  \\
      OB\oursacro{} & ResNet-50 & ImageNet-1k & 5-6 min \\
     
    \bottomrule
  \end{tabular}
\end{table}

\end{document}